%% file: acL_latex.tex
\documentclass[11pt]{article}

\usepackage[final]{acl}

\usepackage{times}
\usepackage{latexsym}

\usepackage[T1]{fontenc}

\usepackage[utf8]{inputenc}

\usepackage{microtype}

\usepackage{inconsolata}

\usepackage{graphicx}
\usepackage{graphicx}
\usepackage[size=tiny]{todonotes}
\usepackage{booktabs}   
\usepackage{makecell}   
\usepackage{adjustbox}  
\usepackage{graphicx}   
\usepackage{tcolorbox}
\usepackage{enumitem}    
\usepackage{placeins} 
\usepackage[table]{xcolor} 
 
\definecolor{darkgreen}{rgb}{0.0, 0.5, 0.0}
\definecolor{rowgray}{gray}{0.95}
\usepackage{float}
\usepackage{subcaption}
\usepackage{tabularx}
\usepackage{array}
\usepackage{amsmath}
\usepackage{placeins}
\usepackage{float}

\usepackage[size=tiny]{todonotes}

\title{When Writing Style Drifts: Benchmarking Authorship Verification under Distribution Shifts in Genre, Time and the AI-Era}
\author{
Lotta Kiefer\textsuperscript{1},
Brisca Balthes\textsuperscript{2},
Christoph Leiter\textsuperscript{3},
Yamen Ajjour\textsuperscript{4},
Elena Schmidt,\\
\textbf{Steffen Eger\textsuperscript{5}}\\
\textsuperscript{1,2,3,4,5}University of Technology Nuremberg (UTN) \\
{\small \textsuperscript{1}lotta.kiefer@utn.de \;
\textsuperscript{5}steffen.eger@utn.de}
}

\begin{document}
\maketitle
\begin{abstract}
Authorship verification (AV) assumes that an author's writing style remains sufficiently stable to distinguish it from that of other writers. In practice, however, this assumption is challenged by distribution shifts caused by changes in genre, time, and AI-assisted writing. Existing AV benchmarks typically study these factors in isolation and focus predominantly on English, limiting our understanding of model robustness under realistic conditions. We introduce AVShift, the first German benchmark for systematically evaluating AV under multiple distribution shifts. AVShift comprises over 150,000 text pairs spanning three genres and 21 years, enabling controlled evaluation of cross-genre, temporal, and AI-era shifts within a unified framework. We benchmark representative feature-based, embedding-based, and LLM-based approaches. Our experiments show that fine-tuned LLMs generalize best across genres and benefit substantially from stylistically diverse training data. We further demonstrate that temporal drift is one of the strongest factors affecting AV, with performance degrading significantly as the time gap between documents increases. In contrast, we find no evidence of a measurable AI-era distribution shift within AVShift. Finally, our feature analysis reveals stylistic features that remain stable across genres, while their relative importance varies depending on the specific genre transition. We release our code and information about dataset access for future research on \href{https://github.com/NL2G/AVShift}{GitHub}.
\end{abstract}

\input{chapters/1_introduction}
\input{chapters/2_related}
\input{chapters/3_dataset}

\input{chapters/4_methods}
\input{chapters/5_results}
\input{chapters/6_discussion}
\input{chapters/7_limitations}
\input{chapters/8_ethics}
\input{chapters/9_acknowledgements}
\bibliography{custom}

\appendix

\input{chapters/10_appendix}

\end{document}

%% file: chapters/1_introduction.tex
\section{Introduction}
\label{sec:intro}
\begin{figure}
    \centering
    \includegraphics[width=\linewidth]{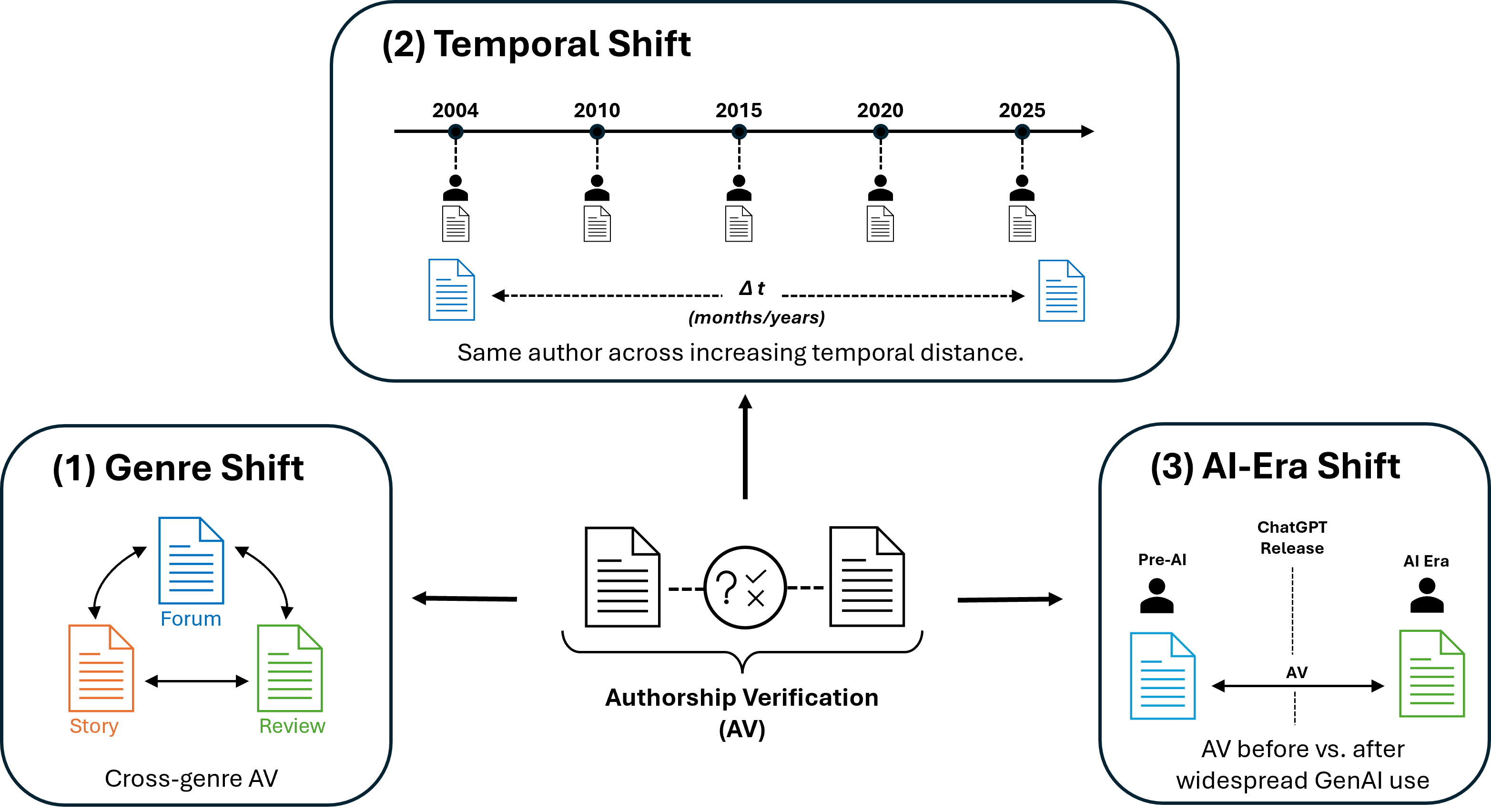}
    \caption{Overview of the stylistic distribution shifts covered by AVShift: cross-genre variation, temporal evolution, and AI-era changes before and after the widespread adoption of generative AI.}
    \label{fig:avshift_overview}
\end{figure}

Authorship analysis aims to identify the author of a text based on characteristic patterns of individual language use, commonly referred to as an \emph{idiolect} \citep{coulthard_idiolect_2004, Nini_2023}. Numerous benchmark datasets spanning books, blogs, news, emails, reviews, social media, and darknet forums have established authorship analysis as an important tool in literary studies, plagiarism detection, and forensic linguistics \citep{tyo2023valla,boenninghoff-etal-2024-wrote,lewis_news_benchmark,manolache_veridark,overdorf_blogs_twitter, Stamatatos2023OverviewOT}.

This work focuses on authorship verification (AV), which determines whether two texts were written by the same author. Unlike authorship attribution that relies on a predefined set of candidates, AV naturally generalizes to previously unseen authors without requiring retraining, making it particularly attractive for forensic investigations involving new suspects.

The concept of linguistic individuality is sometimes described as a \textit{linguistic fingerprint} \citep{kreuz2023linguistic, eder2011stylefingerprints}. However, this analogy can be misleading, giving the wrong impression that writing style is an immutable concept like a biological fingerprint \cite{coulthard_idiolect_2004, Nini_2023}. Instead, linguistic style is influenced by numerous contextual factors, including the intended audience, topic, communicative situation, and medium, and evolves over time.

We therefore view AV as a learning problem that is inherently exposed to distribution shifts. In practice, both training and test data, as well as the paired texts themselves, may differ in genre, time, or broader language use. Existing benchmarks typically investigate these challenges in isolation and focus predominantly on English. However, AV relies on language-specific stylistic cues, making it unclear whether findings obtained on English generalize to other languages. This lack of diverse benchmarks limits our understanding of AV robustness under realistic conditions and across languages.

To address this gap, we introduce \textbf{AVShift}, the first German benchmark for systematically evaluating AV under three key distribution shifts: (1) cross-genre shift across forum posts, reviews, and fanfiction stories, (2) temporal shifts spanning more than two decades of writing, and (3) AI-era shifts before and after the widespread adoption of generative AI (genAI) writing assistants. Figure~\ref{fig:avshift_overview} provides an overview of these evaluation settings.

We make the following contributions:
\begin{itemize}[leftmargin=*, itemsep=0pt, topsep=2pt]
    \item We introduce AVShift, the first German benchmark comprising more than 150,000 text pairs for evaluating AV under cross-genre, temporal, and AI-era distribution shifts.
    \item We provide a systematic evaluation of feature-based, embedding-based, and LLM-based approaches across these distribution shifts. Our results reveal substantial genre-dependent performance differences and show strong cross-genre performance of LLMs trained on mixed-domain data, achieving F1 scores of up to 0.77. While temporal shifts cause performance drops of up to 0.21 F1, we observe no significant performance degradation in the AI-era.
    \item We propose a feature stability score to quantify the robustness of stylistic features across genres and show that feature stability strongly depends on the genre pair.
\end{itemize}

%% file: chapters/2_related.tex
\section{Related Work}
\label{sec:related}

\paragraph{Authorship Analysis Methods}
AV methods can broadly be categorized into feature-based, embedding-based, and LLM-based approaches. 

Feature-based methods represent documents using handcrafted stylistic features, which are compared using similarity measures or statistical or neural classifiers. While these approaches remain attractive due to their competitive performance, efficiency, and interpretability \citep{stamatatos2009survey, grieve2007quantitative}, they have been shown to underperform modern neural methods \cite{zeng2025residualized}.

Embedding-based approaches learn dense stylistic representations directly from text, either through task-specific neural architectures \citep{gupta2019authorship, qian2017deep, boenninghoff2019explainable}  or extracted from pretrained transformer models \citep{rivera2021learning, Ma_Le_Kang_Dou_Cadigan_Freitag_Ritter_Xu_2025, fabien2020bertaa}.

More recently, LLMs have emerged as a promising approach to authorship analysis by jointly learning stylistic representations and the verification task. Although zero- and few-shot prompting with closed-source models such as GPT-3.5 \citep{openai_gpt-3.5} and GPT-4 \citep{openai_gpt-4} achieves competitive performance, their reliance on online APIs limits deployment in forensic and privacy-sensitive settings \citep{huang2024can, ramnath2025cave}. More recent work shows that fine-tuned open-source LLMs outperform both prompting-based approaches and previous AV methods while remaining practical to deploy \citep{hu2024instructav,ramnath2025cave, kiefer-etal-2026-gerav}.

\paragraph{Out-of-Distribution Evaluation}
\textbf{Cross-domain AV} encompasses evaluation settings in which data are drawn from different distributions. This may refer either to domain transfer between training and test data \citep{rivera2021learning} or to the more challenging setting, where the two compared texts originate from different topics, genres, or platforms \citep{overdorf_blogs_twitter, Ma_Le_Kang_Dou_Cadigan_Freitag_Ritter_Xu_2025}. While cross-topic AV has received considerable attention, with numerous methods proposed to reduce topic bias \citep{stamatatos_masking_topic, halvani_posnoise,  hu2023tdrlm5}, cross-platform and cross-genre settings remain comparatively underexplored. Existing work consistently reports substantial performance degradation under these shifts 
\citep{stamatatos2022overview, Stamatatos2023OverviewOT, barlas_cross_domain, israeli-etal-2025-million, Ma_Le_Kang_Dou_Cadigan_Freitag_Ritter_Xu_2025}. Recent work has explored domain-adaptive style representations \citep{Zhang2021ImprovingAV} and training with hard negative and cross-domain positive pairs \citep{VANLEEUWEN2026100943}, improving robustness without eliminating the performance degradation caused by certain domain shifts.

\textbf{Temporal shift} refers to changes in an author's writing style over time, raising the question of how well systems can recognize authors across substantial time gaps. Despite its practical relevance, temporal effects in AV have received limited attention. An existing study on six French novelists shows that performance can degrade as temporal distance increases \citep{cafiero_too_old}. However, the small number of authors limits generalizability, and it remains unclear whether findings from literary texts transfer to other domains, where stylistic choices may be less deliberate. Approaches addressing temporal variation are similarly scarce. \citet{Yang_Zhu_Tang_Wang_2017_topic_drift} model changes in authors' interests over time through topic drift, while \citet{azarbonyad2015time} estimate temporal changes in lexical style and show improvements for authorship attribution on tweets and emails.

\textbf{AI-era shift} refers to stylistic changes introduced by genAI writing assistance. Initial work suggests that AI-assisted writing can increase stylistic similarity between authors and thereby affect verification performance, particularly by increasing false positives \citep{richburg-etal-2024-ai-shift}. This concern is further supported by recent evidence that the widespread use of LLMs as writing assistants reduces linguistic diversity and contributes to stylistic homogenization \citep{sourati_shrinking_2026}. Studying AI shift remains challenging due to the diverse forms of human-AI collaboration, including text generation, revision, and multi-turn interaction \cite{mysore-etal-2025-prototypical, lee2022coauthor}. Many benchmarks disregard the possible influence arising from documents sampled from the AI-era.

\paragraph{Multilingual Evaluation}
Despite the language-dependent nature of stylistic features, authorship analysis remains heavily focused on English. Recent work has introduced multilingual embedding models capable of transferring across languages \citep{qiu-etal-2025-mstyledistance, kim-etal-2025-msr}, and several multilingual benchmarks \citep{israeli-etal-2025-million, murauer2019generating, halvani2016authorship}. German-specific benchmarks have also become available \citep{boenninghoff-etal-2024-wrote, kiefer-etal-2026-gerav}, but these primarily focus on in-domain or cross-topic evaluation. To our knowledge, no benchmark systematically combines different distribution shifts in a multilingual setting.

Overall, prior work has investigated individual distribution shifts largely in isolation and predominantly on English datasets. AVShift addresses this gap by providing the first German benchmark that systematically evaluates AV under cross-genre, temporal, and AI-era distribution shifts within a unified evaluation framework.

%% file: chapters/3_dataset.tex
\section{Data Curation}
\label{sec:data_curation}

The goal of the data collection process is to construct a German corpus for systematically evaluating AV under distribution shifts. After evaluating several candidate sources, we selected \url{www.fanfiktion.de}, which offers three complementary writing environments within a single platform: fanfiction stories, reviews, and forum posts. This allows us to study distribution shifts while minimizing platform-specific confounds.

We use the forum section, specifically \textit{Allgemeines Geplauder} (``General Chit Chat''), as the entry point of our scraping pipeline. Using BeautifulSoup \citep{beautifulsoup}, we extract author profile links from forum threads and subsequently collect all available forum posts, reviews, and fanfiction stories for each author. This produces three aligned corpora containing texts from the same authors across two to three genres.

To capture temporal variation, we traverse the forum archive back to its earliest available entries from 2004. The resulting corpus spans 21 years (2004–2025), enabling both long-term temporal analyses and comparisons between texts written before and after the public release of ChatGPT \citep{openai_gpt-3.5} in November 2022.

\paragraph{Preprocessing}
We remove HTML tags, normalize whitespace, and replace all URLs with a shared token. German message closings are removed together with subsequent content, and remaining self-identifying information is removed through fuzzy username matching.

Texts shorter than 50 words are discarded to ensure sufficient stylistic content, while texts longer than 3,000 words are truncated to reduce computational cost. Unlike previous work, we deliberately retain topical vocabulary and instead control for topic-bias effects during pair construction. Since the website exclusively hosts German-language content, no language filtering is required.

\paragraph{AVShift Benchmark}
AVShift consists of three sub-benchmarks that evaluate complementary distribution shifts: GenreShift, TimeShift, and AIShift. We provide an example in Appendix \ref{sec:av_shift_example}.

\textbf{GenreShift} evaluates cross-genre generalization using seven AV datasets: three in-domain datasets (Forum, Review, Story), three cross-genre datasets (Review-Forum, Story-Forum, Review–Story), and one Mixed dataset obtained by uniformly sampling from the remaining six datasets.

Authors are split uniformly into 80\% training, 10\% validation, and 10\% test partitions across all datasets, ensuring that no author appears in multiple splits, allowing cross-dataset comparison.

Each story sample consists of a single chapter, and each review and forum sample consists of one respective post. To reduce topic leakage, positive pairs are sampled under genre-specific constraints: story pairs originate from different fanfiction works, review pairs review different stories, and forum pairs come from different discussion threads. Positive pairs contain texts by the same author, while negative pairs contain texts by different authors. All datasets are class-balanced, and we sample at most five positive and five negative pairs per author to prevent highly active users from dominating the benchmark. Table \ref{table:data_stat_comp} summarizes the resulting GenreShift datasets.

\begin{table}[t]
\centering
\begin{adjustbox}{width=1\linewidth}
\begin{tabular}{l|rrrr} 
\toprule
 \makecell{\textbf{Dataset}} & \makecell{\textbf{Sample Number}} & \makecell{\textbf{Unique Posts}} & \makecell{\textbf{Unique Users}} & \makecell{\textbf{Mean Sample} \\ \textbf{Len (in Words)}} \\ 
\midrule 
Forum & 15,272 & 16,516 & 2,104 & 173 \\
Review & 22,228 & 23,242 & 2,466 & 184 \\ 
Story & 40,320 & 59,007 & 4,627 & 1,358 \\ 
Review-Forum & 18,490 & 18,274 & 1,849 & 179 \\ 
Review-Story & 26,090 & 35,922 & 2,609 & 786\\
Story-Forum & 32,580 & 37,233 & 3,258 & 787 \\
Mixed & 30,000 & 43,156 & 4,806 & 576 \\
\bottomrule
\end{tabular}
\end{adjustbox}
\caption{Statistical Comparison: All GenreShift datasets}
\label{table:data_stat_comp}
\end{table}

\textbf{TimeShift} evaluates robustness to temporal shift. We construct ten dataset slices covering temporal gaps of 12 months each, ranging from 0-12 months to 108-120 months. All text pairs are sampled such that the publication dates of the two texts in each pair fall within the corresponding temporal interval of a given slice (e.g., in the 12-24 month slice, the publication dates of Text A and Text B differ by at least 12 and less than 24 months in both positive and negative pairs). This procedure is applied independently to the Forum, Review, and Story corpora. Pairs are sampled independently of the original test split, using authors from the corresponding genre corpus to ensure sufficient samples under each temporal condition. To ensure comparability, all datasets are downsampled to the size of the smallest subset within each genre (Forum: 670 pairs, Review: 446 pairs, Story: 4,466 pairs).

\textbf{AIShift} evaluates the impact of the widespread adoption of genAI on AV. The corpus is partitioned into four periods: \textit{Early (2004–2010)}, \textit{Mid (2011–2016)}, \textit{Pre-AI (2017–2022)}, and \textit{AI (2023–2025)}. To ensure no additional effect of the varying time spans of the different era subsets, we restrict the time difference between texts within one text pair to $\leq 3$ years, matching the AI-era condition (2023-2025) with the shortest time difference. Following the same pair construction procedure as GenreShift, we construct separate datasets for each genre and period, enabling controlled comparisons of AV before and after the emergence of AI-assisted writing. As for TimeShift, these pairs are generated independently of the original dataset splits.

%% file: chapters/4_methods.tex
\section{Experimental Setup}
\label{sec:experimental_setup}
To evaluate AV under different distribution shifts, we compare representative approaches covering the dominant AV paradigms: feature-based, embedding-based, and LLM-based methods.

\paragraph{AV Models}
As a representative feature-based approach, we train an XGBoost classifier \citep{xgb_chen_2016} on handcrafted stylometric features. We extract a comprehensive set of more than 4,000 distinct stylistic features (see Appendix \ref{sec:xgb_features} for details). Feature vectors are extracted independently for both texts and combined using their element-wise difference before classification. Additionally, we evaluate an n-gram-based approach following \citet{weerasinghe2021feature}, which represents texts as character n-gram feature vectors and uses a logistic regression classifier for AV.

As embedding-based methods, we evaluate three approaches. First, we use the Multilingual Style Representation (MSR) model proposed by \citet{kim-etal-2025-msr}, which learns language-agnostic stylistic embeddings from 36 languages and 13 domains and achieves strong performance, even on unseen languages such as German. Second, we include mStyleDistance \citep{qiu-etal-2025-mstyledistance}, a style embedding model trained on multilingual data that includes German. For both models, AV is performed using the cosine similarity between the embeddings of the two texts, with decision thresholds calibrated on the validation set. Third, we evaluate a Siamese network based on BERT, which learns text representations through contrastive training \citep{fabien2020bertaa}. Following \citet{kiefer-etal-2026-gerav}, we adapt this approach to German AV by using the German BERT model \citep{german-bert}.

For the LLM-based approach, we follow the framework of \citet{kiefer-etal-2026-gerav}, which fine-tunes instruction-tuned LLMs with LoRA \citep{hu2022lora} to answer the binary question of whether two texts were written by the same author. We replace their best-performing model, Gemma-3-12B-it \citep{gemmateam2025gemma3technicalreport}, with the more recent Gemma-4-31B-it \citep{gemmateam2026gemma4technicalreport}, while keeping the training procedure unchanged. In addition, we evaluate a prompting-based approach using the LIP prompt component \citep{huang2024can} with the same Gemma model, which instructs the model to first analyze the stylistic characteristics of both texts before making an authorship decision.

\paragraph{Evaluation Metrics}
We report binary F1 scores for the positive (same-author) class, following prior AV literature \citep{ kestemont2020overview, stamatatos2022overview, Ma_Le_Kang_Dou_Cadigan_Freitag_Ritter_Xu_2025}. We additionally report Accuracy scores in the Appendix.

\paragraph{GenreShift Evaluation}
We train and evaluate all seven models on each of the seven AVShift datasets (see Appendix \ref{sec:hardware_hyperparmaters} for the training setup). Note that for MSR and mStyleDistance, the embeddings remain unchanged, and only the decision threshold is tuned on AVShift using Youden’s $J$ statistic \citep{youden1950index}. LIP prompting is performed zero-shot using Gemma. This evaluation setup assesses both generalization to unseen domains and within-sample cross-genre AV performance. Statistical significance of performance differences between models is assessed using paired bootstrap testing with 10,000 resamples \citep{efron_bootstrap_1979}.

Because AV performance is strongly influenced by text length and can be affected by training set size \citep{eder2015size, kiefer-etal-2026-gerav}, we additionally construct standardized GenreShift datasets. Every document is normalized to 500 words, extending shorter texts by concatenating additional texts from the same author and truncating longer texts. We further downsample all datasets to the size of the smallest genre split (3,640 samples), allowing us to isolate the effect of genre independently of text length and dataset size. The standardized GenreShift datasets are marked by an additional \textit{500}.

To further investigate stylistic variation across genres, we exploit the interpretability of the handcrafted feature representation used in the XGB approach. We first concatenate all texts written by each author into a single document and balance the three resulting genre corpora by truncating them to the same total token count. A shared feature vectorizer is then fitted on the combined author corpora of stories, reviews, and forum posts, enabling direct comparison of feature distributions across genres. This analysis allows us to identify stylistic features that remain stable across genres as well as those that are strongly genre-dependent. 

\paragraph{TimeShift Evaluation}
For each genre, we evaluate the best-performing in-domain model on the TimeShift benchmark. Performance is measured on ten datasets with temporal gaps ranging from 0–12 months to 108–120 months. We report Pearson \citep{Pearson1895} and Spearman \citep{spearman1904general} correlation coefficients between temporal distance and verification performance to quantify the effect of temporal drift. 

\paragraph{AIShift Evaluation}
We evaluate AIShift using a leave-one-era-out protocol. For each experiment, one temporal period is held out for testing while the remaining three periods are combined for training. For each split, we uniformly sample the same number of pairs (train: 4,122, test: 1,374) to ensure comparable dataset sizes across all experiments. This setup enables a controlled evaluation of AV performance before and after the emergence of AI-assisted writing across different genres.

\paragraph{CrossNews}
To assess the generalizability of our findings beyond German, we additionally evaluate our methods on the English CrossNews benchmark introduced by \citet{Ma_Le_Kang_Dou_Cadigan_Freitag_Ritter_Xu_2025}. CrossNews links news articles and tweets written by the same author, yielding two in-domain datasets (Article and Tweet) and one cross-genre dataset (Article-Tweet). Including an English benchmark allows us to examine whether the distribution-shift effects observed on AVShift extend beyond German and to identify potential language-specific differences. We compare our methods against the two best-performing approaches reported by the authors: (1) a prompting-based method using LLaMA-3-70B \cite{grattafiori2024llama3herdmodels}, and (2) SELMA, which performs AV using embedding distances obtained with e5-mistral-7b-instruct \citep{wang-etal-2024-improving-text-mistral}.

%% file: chapters/5_results.tex
\section{Results}
\label{sec:results}
\begin{figure}[h!]
    \centering
    \includegraphics[width=\linewidth]{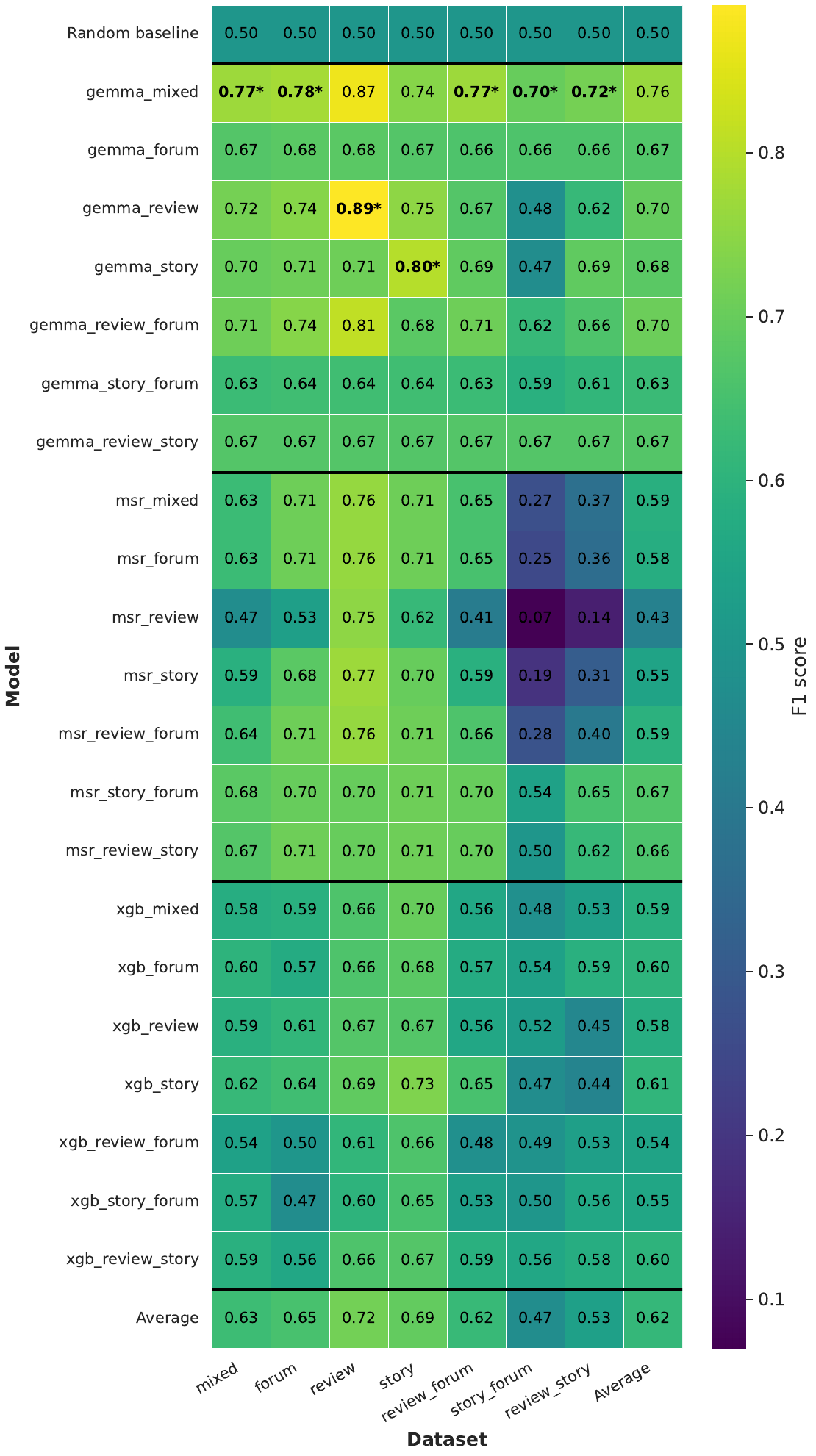}
    \caption{F1 scores of the three representative models trained and evaluated on all GenreShift train and test splits. Model names are
shown on the y-axis and test dataset names on the x-axis. Each model name is followed by its training or calibration dataset. The best score for each test set is shown in bold. Statistically significant superiority over all other models is indicated by an asterisk (*; p < 0.05).}
    \label{fig:genre_shift}
\end{figure}
\paragraph{How robust are models across genres?}
Our results on the GenreShift benchmark are summarized in Figure~\ref{fig:genre_shift}. For clarity, we show only the best-performing model from each methodological family across the majority of test datasets: LoRA-tuned Gemma for LLM-based approaches, XGB for feature-based approaches, and MSR for embedding-based approaches. Full results for all evaluated models, including accuracy scores, are provided in Appendix \ref{sec:full_results_genre}. Models are shown on the y-axis, followed by their respective training set names, while test datasets are shown on the x-axis. In-domain datasets are denoted by their genre name (e.g., Review), whereas cross-genre datasets are denoted by both genres from which the text pairs are drawn (e.g., Review-Forum, where a pair consists of one text from Review and one text from Forum). Across all seven datasets Gemma consistently outperforms both the feature-based XGB classifier and the embedding-based MSR model, achieving the highest F1 score on every test set ($p<0.05$). While models trained on the Review and Story datasets perform best on their respective in-domain tasks, the Mixed Gemma model achieves the strongest performance on all remaining datasets. This demonstrates that exposing LLMs to stylistically diverse training data substantially improves robustness to distribution shifts.

Among the three in-domain datasets, Review consistently proves to be the easiest genre for AV, followed by Story and Forum. The best-performing models evaluated on the in-domain datasets achieve F1 scores of 0.89, 0.80, and 0.78, respectively. This suggests that reviews contain the strongest and most consistent authorial signal, whereas forum posts represent the most challenging writing style.

Both Gemma and XGB degrade substantially under cross-genre transfer. For example, Gemma achieves an F1 score of 0.89 when trained and tested on reviews but loses approximately 0.2 F1 when trained on stories or forum posts. The Forum dataset is an exception, where training on reviews or stories yields better performance than training on forum data, suggesting that forum posts provide weaker supervision for learning robust stylistic representations. In contrast, MSR is less sensitive to the calibration genre, likely because its embedding model remains fixed and only the verification threshold is adapted to AVShift.

 The three cross-genre datasets also differ substantially in difficulty. Review-Forum is consistently the easiest cross-genre setting, whereas Review–Story and Story-Forum are considerably more challenging. Surprisingly, the Mixed Gemma model outperforms models trained directly on the corresponding cross-genre datasets in every setting. This indicates that exposing the model to a broad range of stylistic variation is more beneficial than specializing on a single genre transition. Notably, its performance on Review-Forum approaches the in-domain performance obtained on the Forum dataset (0.77 F1), demonstrating that robust cross-genre AV is achievable when sufficient stylistic diversity is observed during training. While previous work consistently reported substantial performance degradation under cross-genre evaluation \citep{Ma_Le_Kang_Dou_Cadigan_Freitag_Ritter_Xu_2025,israeli-etal-2025-million, Stamatatos2023OverviewOT}, our results suggest that much of this degradation can be mitigated through sufficiently diverse training data. 

 To determine whether these differences are genuinely caused by genre rather than confounding factors such as text length or training set size, we repeat the experiments on our standardized AVShift datasets in which both factors are controlled. We limit this experiment to the best-performing models of each methodological family as reported before (XGB, MSR and LoRA-tuned Gemma). The relative difficulty of the three genres remains unchanged: Review (0.86 F1) continues to outperform Story and Forum (both 0.74 F1) (see Appendix \ref{sec:full_results_genre_stand}), confirming that the observed ranking is intrinsic to the writing genres rather than an artifact of dataset construction. In contrast, the ranking of the models changes considerably. Under these controlled conditions, Gemma is no longer consistently superior. MSR achieves the highest performance on the standardized Review and Story datasets, while XGB performs best on Forum, although the best-performing Gemma models follow closely and do not differ significantly ($p\geq0.05$). At the same time, cross-genre performance deteriorates for all models, which we attribute primarily to the substantially smaller training sets resulting from the controlled sampling procedure. Together, these findings indicate that genre itself is the dominant source of difficulty in AVShift, while the relative performance of different AV approaches depends strongly on the characteristics of the training data. In particular, XGB and MSR benefit from the standardized setting and appear less sensitive to the reduced training size, whereas Gemma benefits more from the larger and stylistically more diverse training data available in the original benchmark.

\paragraph{Do our results generalize to English?}
To assess whether the trends observed on AVShift generalize beyond German, we evaluate all three approaches on the English CrossNews benchmark. Table \ref{table:crossnews} compares the best-performing models reported by \citet{Ma_Le_Kang_Dou_Cadigan_Freitag_Ritter_Xu_2025} with the strongest model results we achieved on their benchmark. Overall, our results closely mirror the findings on AVShift. Gemma achieves the best performance on the Tweet–Tweet and Article–Tweet datasets, improving upon the previously reported state-of-the-art by 0.11 and 0.07 F1, respectively. On the Article–Article dataset, MSR achieves the highest performance, slightly outperforming SELMA. These results demonstrate that the strong cross-genre generalization of Gemma is not limited to German but also transfers to English.

Overall, the similar trends across CrossNews and AVShift suggest that the observed distribution-shift effects are not exclusive to German. However, this does not imply that language-specific evaluation is redundant: AV performance can depend substantially on the target language, and similar trends do not necessarily translate into identical model behavior or performance. Previous work has shown that training on the target language is essential for high performance \citep{kiefer-etal-2026-gerav}. AVShift therefore complements existing English resources by enabling the systematic investigation of distribution shifts in German.

CrossNews remains consistently easier than AVShift, with substantially higher F1 scores across datasets. While this may partly reflect the predominantly English pretraining of models such as Gemma, the benchmarks differ in several other aspects, preventing attribution of the performance gap to language alone.

\begin{table}
\centering
\begin{adjustbox}{width=01\linewidth}
\begin{tabular}{lrrr} 
\toprule
 \makecell{\textbf{Model}} & \makecell{\textbf{Article-Article}} & \makecell{\textbf{Tweet-Tweet}} & \makecell{\textbf{Article-Tweet}}\\ 
\midrule 
\citet{Ma_Le_Kang_Dou_Cadigan_Freitag_Ritter_Xu_2025} LLaMA Prompting & 0.77{\scriptsize$\pm0.089$} & 0.79{\scriptsize$\pm0.048$} & 0.40{\scriptsize$\pm0.064$} \\
\citet{Ma_Le_Kang_Dou_Cadigan_Freitag_Ritter_Xu_2025} SELMA
 & 0.86{\scriptsize$\pm0.018$} & 0.75{\scriptsize$\pm0.020$} & 0.80{\scriptsize$\pm0.023$} \\ \midrule
Gemma-AT & 0.83 & 0.88 & \textbf{0.87} \\ 
Gemma-TT & 0.83 & \textbf{0.90} & 0.86 \\
MSR-AA & \textbf{0.88} & 0.84 & 0.69 \\
\bottomrule
\end{tabular}
\end{adjustbox}
\caption{Results on the CrossNews benchmark. The two models scoring best for the three benchmark subsets as reported by \citet{Ma_Le_Kang_Dou_Cadigan_Freitag_Ritter_Xu_2025} are shown alongside the best-scoring models from our analysis.}
\label{table:crossnews}
\end{table}

\begin{figure}
    \centering
    \includegraphics[width=\linewidth]{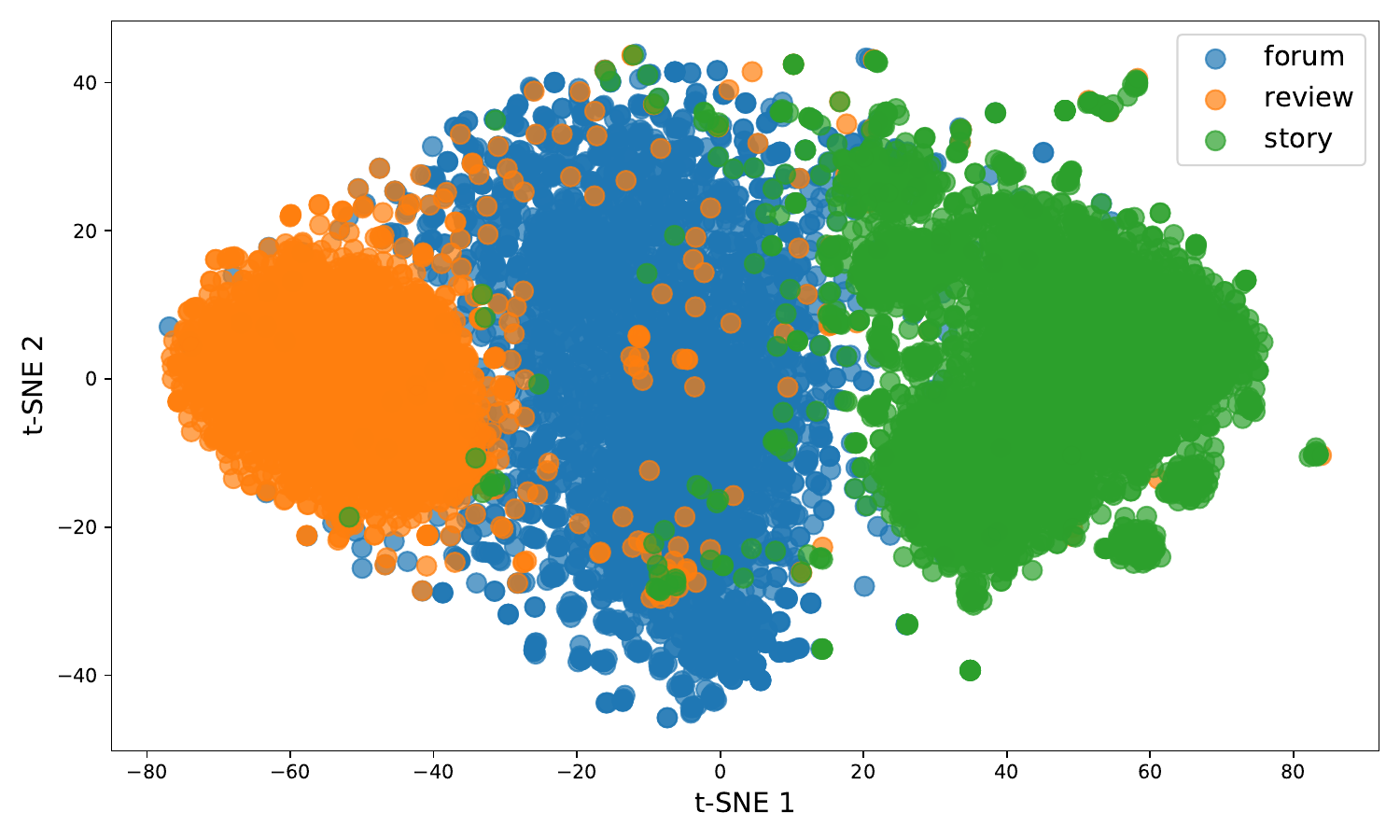}
    \caption{t-SNE visualization of feature author vectors from different genres.}
    \label{fig:tsne}
\end{figure}

\paragraph{Do Stylistic Features Survive Genre Shifts?}
To better understand why some genre transitions are more challenging than others, we analyze the shared handcrafted feature space described in Section \ref{sec:experimental_setup}. Figure~\ref{fig:tsne} shows a t-SNE \citep{cai_tsne} projection of the resulting author representations. Rather than clustering primarily by author, the vectors are largely separated by genre, indicating that genre exerts a strong influence on the handcrafted feature representation even for texts written by the same individual.

To quantify how well individual features preserve authorial style across genres, we compute a cross-genre feature stability score \begin{equation}
 \text{Stability}(f) = 1 - \frac{\sigma_{\mathrm{within}}(f)}{\sigma_{\mathrm{between}}(f) + \varepsilon} \end{equation} which compares within-author variation across genres to between-author variation. High stability indicates that a feature remains consistent for the same author while discriminating between different authors and representing robust indicators of authorial style.

Across all three genres, stability scores range from $-1.70$ to 0.90, with a mean of 0.22 and a median of 0.25 (Table \ref{table:stability_stats}). Overall, 80\% of the handcrafted features exhibit positive stability, indicating that most stylistic features remain relatively consistent across genres despite the strong genre separation observed in the t-SNE projection. This suggests that successful cross-genre AV remains feasible. We provide the thirty most and least stable features in Appendix \ref{sec:feature_ordering}.

Feature stability further reflects differences in cross-genre verification difficulty. Review-Forum exhibits the highest average stability (0.31 and 85\% positive features), consistent with the results showing highest classification performance. Even though Review-Story ranks second in performance it shows the lowest stability scores in our analysis ($-0.06$; 48\% positive features). Thus, feature stability does not perfectly predict verification performance, but the overall trend suggests that it is a useful indicator of how well authorial style is preserved across genres and, consequently, of the expected difficulty of cross-genre AV.

Finally, we investigate whether the same features remain stable across different genre transitions. We look at the overlap among the 100 most stable and 100 least stable features and find that overlap is generally low, ranging from 2\% to 43\%, indicating that different genre transitions affect different subsets of stylistic features. Nevertheless, the overall feature rankings remain moderately correlated (Spearman's $\rho=0.42-0.72$), suggesting that while genre shifts change which features are most informative, the broader ordering of feature stability is largely preserved (see Appendix \ref{sec:pairwise_feature_analysis}). Taken together, these findings indicate that feature stability should be analyzed separately for each genre transition rather than assuming a universal set of robust stylistic features. 

\begin{table}[h]
\centering
\scalebox{0.9}{
\begin{adjustbox}{width=1\linewidth}
\begin{tabular}{l|rrr} 
\toprule
 \makecell{\textbf{Dataset}} &\makecell{\textbf{Mean Stability}} & \makecell{\textbf{Median Stability}} & \makecell{\textbf{Percentage Stable} \\ \textbf{Features}} \\ 
\midrule 
Overall & 0.22{\scriptsize$\pm0.28$} & $0.25$ & 80\% \\
Review-Forum & 0.31{\scriptsize$\pm0.27$} & $0.34$ & 85\% \\
Story-Forum & 0.19{\scriptsize$\pm0.37$} & $0.25$ & 75\% \\ 
Review-Story & $-0.06${\scriptsize$\pm0.38$} & $-0.01$ & 48\% \\ 
\bottomrule
\end{tabular}
\end{adjustbox}
}
\caption{Mean and median stability feature scores alongside the percentage of features with positive score for the whole dataset next to each genre-pair individually.}
\label{table:stability_stats}
\end{table}

\paragraph{How does authorial style change over time?}
\begin{figure}[!h]
    \centering
    \includegraphics[width=\linewidth]{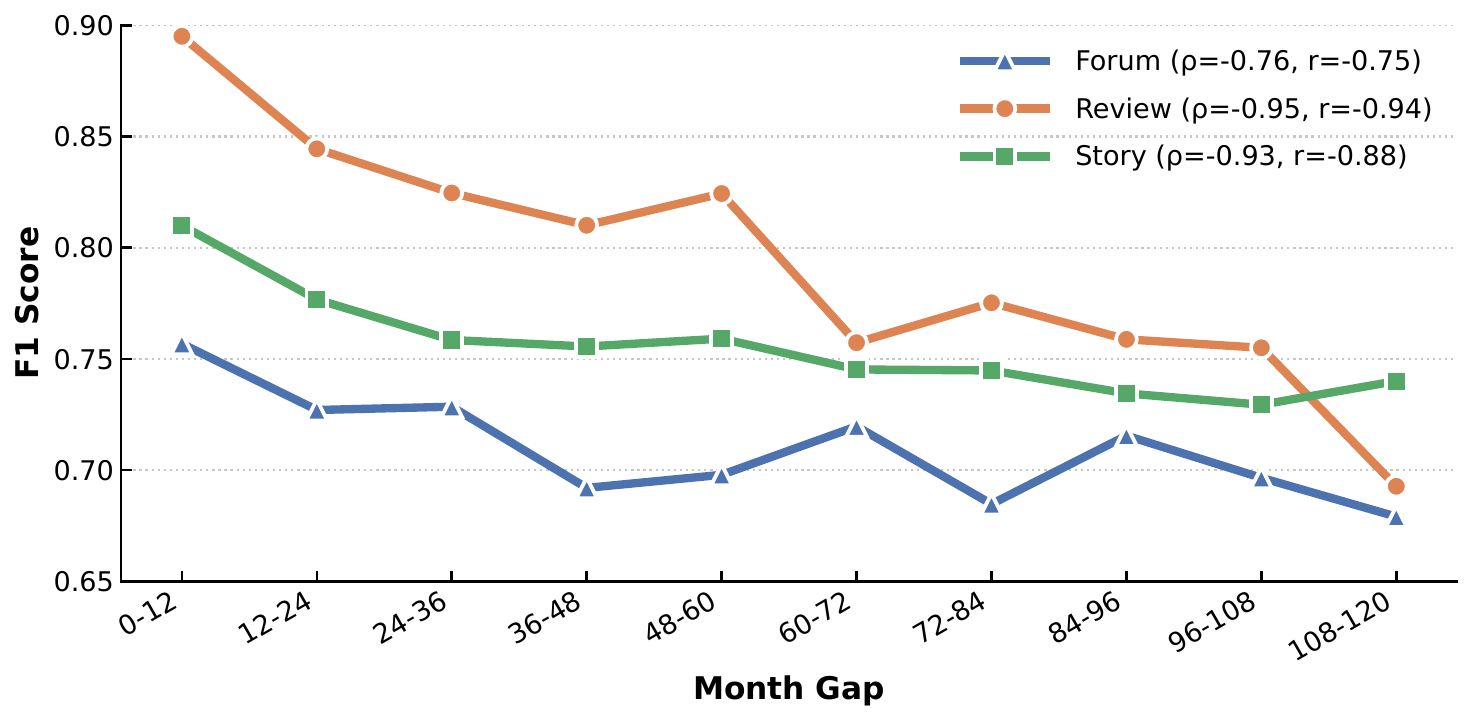}
    \caption {Correlation of increasing time spans (x-axis) and F1 score (y-axis) alongside Pearson (r) and Spearman ($\rho$) correlation coefficients}
    \label{fig:time_plot}
\end{figure}

Figure~\ref{fig:time_plot} shows the performance of the best-performing in-domain model for each genre on the TimeShift benchmark (the Mixed Gemma model for Forum and the respective genre-specific fine-tuned Gemma models for Review and Story). Across all genres, AV performance decreases steadily as the temporal gap between two texts increases. Pearson and Spearman correlation analyses reveal a strong and statistically significant negative relationship between temporal distance and F1 score for all genres ($p<0.05$).

The magnitude of this degradation differs across genres. Review shows the largest decline, with the F1 score dropping from 0.90 for text pairs separated by 0–12 months to 0.69 after 9–10 years. In contrast, Story and Forum show smaller decreases ($\sim$0.08 F1), although this may partly reflect their lower initial performance, leaving less room for degradation. Despite these differences, the relative difficulty of the three genres remains largely unchanged (Review > Story > Forum) across temporal intervals, suggesting that the higher performance on Review is unlikely to be attributable to potential differences in the time spans across genres. 

These findings demonstrate that authorial style is not static but evolves continuously over time, substantially reducing verification performance even for state-of-the-art models. Notably, a large decline occurs already after the first year, showing that temporal drift emerges very early. Since all three genres exhibit the same overall trend, temporal variation should be considered an important factor when constructing and evaluating AV benchmarks. For high-stakes real-world applications in particular, our results suggest that verification is only reliable when comparing documents written within relatively short time windows without further model modifications.

\paragraph{Has the emergence of genAI changed AV?}
\begin{figure}[h!]
    \centering
    \includegraphics[width=\linewidth]{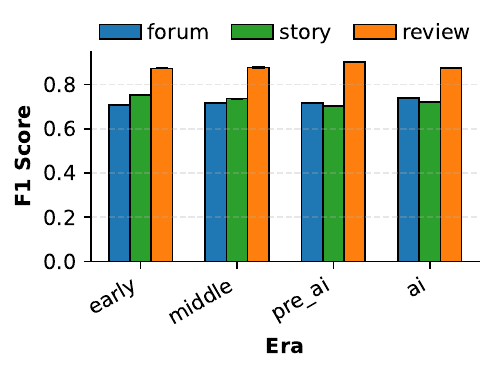}
    \caption{Gemma performance under different era evaluations for each genre. Datasets on the x-axis refer to the held-out test era with F1 scores on the y-axis.}
    \label{fig:era_plot}
\end{figure}

Figure \ref{fig:era_plot} shows the F1 scores for the leave-one-era-out evaluation across all three genres. We find no evidence of systematic performance degradation when evaluating on AI-era texts compared to earlier periods. Although performance differs significantly (p < 0.05) between individual hold-out eras, these differences do not follow a consistent chronological pattern. For example, the Story dataset performs best in the Early era (0.75 F1) and exhibits its largest performance drop when the Pre-AI era is held out (0.70 F1), whereas the Forum dataset achieves its highest performance on the AI-era (0.74 F1). This suggests that performance variation is more likely driven by dataset-specific factors, such as text length, topic, or author composition, than by publication era alone.

These findings should, however, be interpreted with caution, as our corpus lacks annotations of AI-assisted writing, preventing us from determining whether or to what extent texts were generated or modified using genAI. One possible explanation for the absence of performance degradation is that the fanfiction community's reservations towards AI-generated content, as reported by \citet{Alfassi2026fanfiction}, may have discouraged its use within our corpus. Consequently, our results do not establish that AI-assisted writing does not affect AV. Rather, they indicate no consistent performance degradation coinciding with the emergence of genAI writing assistants within our corpus under controlled sampling conditions. Future work should investigate the impact of AI assistance using datasets with annotated or controlled levels of AI use.

%% file: chapters/6_discussion.tex
\section{Conclusion}
\label{sec:conclusion}
In this work, we introduced AVShift, the first German benchmark for systematically evaluating AV under realistic distribution shifts. AVShift unifies cross-genre, temporal, and AI-era evaluation, enabling a comprehensive assessment of robustness beyond conventional in-domain settings. We use it to compare feature-based, embedding-based, and LLM-based approaches under challenging real-world conditions.

Our experiments reveal three main findings. First, although cross-genre AV remains challenging, fine-tuned LLMs perform particularly well, benefiting from stylistically diverse training data and even matching in-domain performance in one setting. Second, temporal drift strongly affects AV, with performance consistently declining as the time gap between documents increases. Third, we find no evidence that the emergence of genAI has introduced a distribution shift in the form of systematic performance degradation on AI-era texts within AVShift. However, as the corpus lacks annotations of AI-assisted writing, the direct impact of genAI assistance on AV remains to be investigated using datasets with controlled AI use.

Beyond benchmarking, our analyses provide new insights into authorial style. Although handcrafted features are strongly influenced by genre, many stylistic features remain stable enough to support reliable cross-genre AV. Together with the superior performance of models trained on stylistically diverse data, these findings suggest that robustness is achieved not by eliminating stylistic variation, but by learning representations that capture author-specific characteristics despite changes in writing context.

We hope AVShift will serve as a valuable resource for robust AV research, particularly for non-English languages where benchmark datasets remain scarce. More broadly, our results suggest that future progress should be measured not only by in-domain performance but also by robustness to realistic distribution shifts, providing a more reliable assessment for forensic and other real-world applications.

%% file: chapters/7_limitations.tex
\section*{Limitations}
This work has several limitations that should be taken into consideration in the interpretation of the results. First, our evaluation covers a representative but limited selection of AV approaches. While we select seven models representing feature-based, embedding-based, and LLM-based methods, other approaches may yield different results. Moreover, within each model category, design choices such as LLM architecture, fine-tuning strategy, threshold calibration, or feature selection may influence absolute performance and model rankings. A broader evaluation across additional methods and configurations remains an important direction for future work.

Second, AVShift is constructed from a single German platform, which enables controlled comparisons across genres while reducing platform-specific confounding factors. However, this also limits the diversity of writing environments represented in the benchmark. Future extensions incorporating additional platforms and languages could further assess the generalizability of the observed findings.

Third, while AVShift introduces realistic distribution shifts, measuring some forms of shift remains challenging. In particular, AI-era shift depends on the extent and type of human-AI interaction, which we did not try to quantify in this work. Future work could investigate controlled settings with known levels of AI assistance.

Finally, this work focuses on analyzing robustness under distribution shifts rather than developing methods to improve style shift robustness. We provide AVShift and our analyses as a foundation for future research on adaptive training strategies, robust representations, and methods specifically designed to address distribution shifts in AV.

%% file: chapters/8_ethics.tex
\section*{Ethical Considerations}
We aim to minimize the environmental impact of our experiments by restricting GPU usage to the resources required for model training and evaluation.

AV has potential societal benefits in applications such as forensic investigations and plagiarism detection. However, we acknowledge that these technologies may also be misused, for example, to deanonymize individuals or undermine legitimate privacy protections. Furthermore, AV systems are inherently imperfect, and their predictions should not be interpreted as definitive evidence in real-world applications. In particular, we cannot fully exclude the influence of demographic, social, or other contextual factors that may introduce biases against specific groups.

We provide details on model and dataset licenses in Appendix \ref{sec:licences}. The dataset will be available through gated access for research purposes only, and usernames will be pseudonymized. The underlying platform provides publicly accessible content without requiring user authentication; nevertheless, we recognize that publicly available data may still carry privacy considerations and encourage responsible use of the resource.

%% file: chapters/9_acknowledgements.tex
\section*{Acknowledgements}
\label{sec:acknowledgements}
We gratefully acknowledge the support that made this work possible. This work was supported by the German Federal Ministry of Research, Technology and Space (BMFTR) through the ALiAS research project (grants 13N17272 and 13N17273) within the security research program, and by the German Research Foundation (DFG) under the Heisenberg Grant EG 375/5-1.

%% file: chapters/10_appendix.tex
\section{AVShift Example}
\label{sec:av_shift_example}
We present document examples from each genre written by a single author in Table \ref{table:text_examples}, together with their English translations. These examples provide an impression of how writing style varies across genres. The forum text is relatively informal, for instance using emojis such as "xD", whereas the story example differs substantially by employing a more literary style. The review is again more informal but has the distinctive characteristic of directly addressing the author of the story.

\begin{table*}[h!]
\centering
\begin{tabularx}{\linewidth}{XXX}
\toprule
\makecell{\textbf{Forum}} & \makecell{\textbf{Story}} & \makecell{\textbf{Review}} \\
\midrule
Im Sommer saß ich mal mit einer Freundin in meinem Zimmer am Fußboden. Wir haben gemalt, da fiel auf einmal eine riesige Raupe/Larve von der Decke! Wir wissen bis heute nicht, wie sie da hin gekommen ist.Das andere ist im Sommer am Schulfest passiert.
Ich saß im Gras, als mir plötzlich ein Vogel was auf's Knie fallen ließ - genau so wie letztes Jahr. xD & Megatron bewegte sich durch
die dunklen Gänge, doch seine Gedanken waren ferner denn je.
Jeder Schritt quälte die Ruhe, wie der Ton eines fallenden
Tropfens die endliche Stille. Seine Präsenz füllte den Ort wie
das Summen einer Stimmgabel, das in jede Ecke drang, sich
selbst in jenen Flächen nieder ließ, die den Raum begrenzten,
um ihn aus seinem Frieden zu reißen. […] & Hi, hui ja, das ist wirklich ein ungewöhnliches Pairing xD Aber nicht schlecht, dein Stil ist eigentlich richtig gut und es lässt sich alles flüssig lesen. Ich schließe mich Hera an und meine, dass Absätze im Text nicht schlecht gewesen wären. Du musst bedenken, dass die Geschichte am Bildschirm gelesen wird, was anstrengend für die Augen ist. […] \\
\midrule
\multicolumn{3}{c}{\textbf{Translation}} \\
\midrule
One summer, I was sitting on the floor in my room with a friend. We were drawing when suddenly a huge caterpillar/larva fell from the ceiling! To this day, we still don't know how it got there. The other thing happened at the school festival that summer.
I was sitting in the grass when suddenly a bird dropped something on my knee - just like last year. xD & Megatron moved through the dark corridors, yet his thoughts were farther away than ever. Each step disturbed the stillness, like the sound of a falling drop breaking the finite silence. His presence filled the place like the hum of a tuning fork, penetrating every corner, settling even into the surfaces that bounded the room, to tear it from its peace. […] & Hi, wow, yeah, that’s really an unusual pairing xD But not bad, your writing style is actually really good, and it all reads smoothly. I agree with Hera - I think some paragraphs in the text wouldn’t have been a bad idea. You have to keep in mind that the story is being read on a screen, which can be hard on the eyes. […] \\
\bottomrule
\end{tabularx}
\caption{Example of texts from one user writing in all three genres alongside English translations. Usernames occurring in the examples have been pseudonymized.}
\label{table:text_examples}
\end{table*}

\section{XGB Feature Configuration}
\label{sec:xgb_features}
Table \ref{table:xgb_features} presents the feature configuration used in the feature-based approach. It lists each feature abbreviation together with a brief description.

\begin{table}[h!]
\centering
\begin{tabular}{l|p{0.58\linewidth}}
\toprule
\textbf{Feature} & \textbf{Description} \\
\midrule
mfw2 & Normalized frequencies of the 1,000 most frequent word bigrams. \\
mft & Normalized frequencies of the 1,000 most frequent POS trigrams. \\
mfc & Normalized frequencies of the 2,500 most frequent character 4-grams. \\
mfe & Normalized frequencies of the 100 most frequent emojis. \\
wordLenDistri & Distribution of word lengths from 1 to 20 characters. \\
wordLen & Average word length. \\
messageLen & Average document length. \\
nrPunctuation & Normalized frequencies of individual punctuation symbols. \\
nrOOV & Proportion of out-of-vocabulary words. \\
\bottomrule
\end{tabular}
\caption{Handcrafted features used to train the feature-based XGB AV model.}
\label{table:xgb_features}
\end{table}

\section{Training Setup and Hyperparameters}
\label{sec:hardware_hyperparmaters}

To support LoRA fine-tuning of the Gemma-4-31B-it model, we conduct our experiments on a system equipped with four H200 GPUs. We follow the hyperparameter configuration proposed by \citet{kiefer-etal-2026-gerav} and update the software stack to support the newer Gemma version. Specifically, we use PyTorch 2.12.1 \citep{Ansel_PyTorch_2_Faster_2024}, Transformers 5.12.1 \citep{Wolf_Transformers_State-of-the-Art_Natural_2020}, TRL 0.21.0 \citep{von_Werra_TRL_Transformers_Reinforcement_2020}, and vLLM 0.24.0 \citep{kwon2023efficient}.

For the feature-based XGBoost model, we use an environment with Transformers 4.36.2 and PyTorch 2.5.1. We perform a dedicated hyperparameter grid search for each training setting, tuning the maximum tree depth (3, 6, 10, 15), minimum child weight (0, 2, 4, 5), regularization parameter $\alpha$ (0, 1), $\gamma$ (0, 1), and the learning rate (0.01, 0.1, 0.3).

The MSR and mStyleDistance models are evaluated using Transformers 5.12.1, Sentence-Transformers 5.2.2, and PyTorch 2.12.1. We use the original sentence embeddings without modification and tune only the verification threshold for each dataset.

We release complete environment and training setups within our GitHub repository for reproducibility.

\section{GenreShift Full Results}
\label{sec:full_results_genre}
\begin{figure}[h!]
    \centering
    \includegraphics[width=\linewidth]{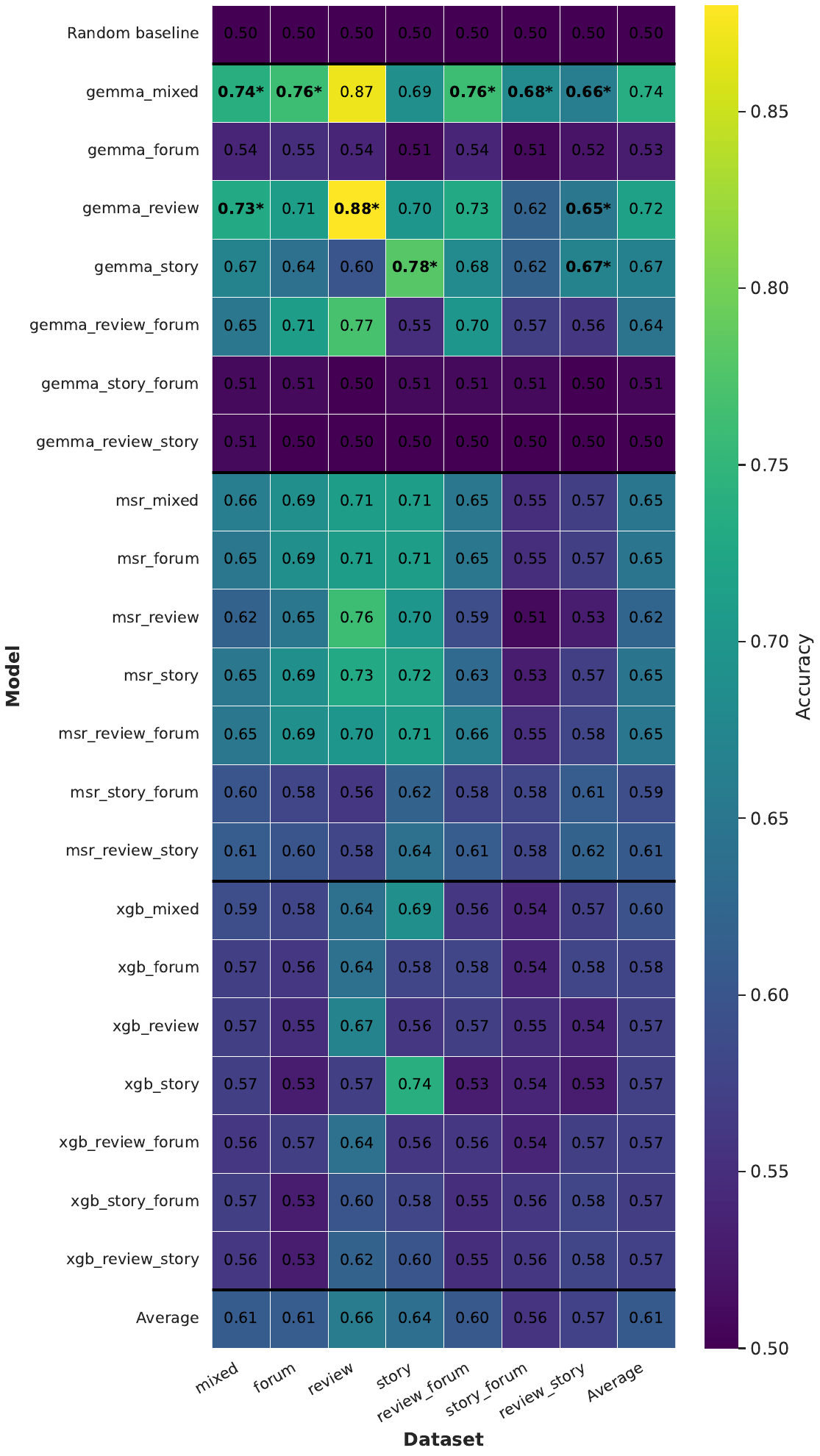}
    \caption{Accuracy scores of the three representative models trained and evaluated on all GenreShift train and test splits. Model names are shown on the y-axis and test dataset names on the x-axis. Each model name is followed by its training or calibration dataset. The best score for each test set is shown in bold. Statistically significant superiority over all other models is indicated by an asterisk (*; p < 0.05).}
    \label{fig:acc_all_names}
\end{figure}

Figure \ref{fig:acc_all_names} reports accuracy scores to complement the F1 results presented in the main results section. The overall findings remain unchanged, with Gemma consistently outperforming the other models and the review genre reaching highest performance.

\begin{figure*}[h!]
    \centering
    \begin{subfigure}[b]{0.48\textwidth}
        \centering
        \includegraphics[width=\linewidth]{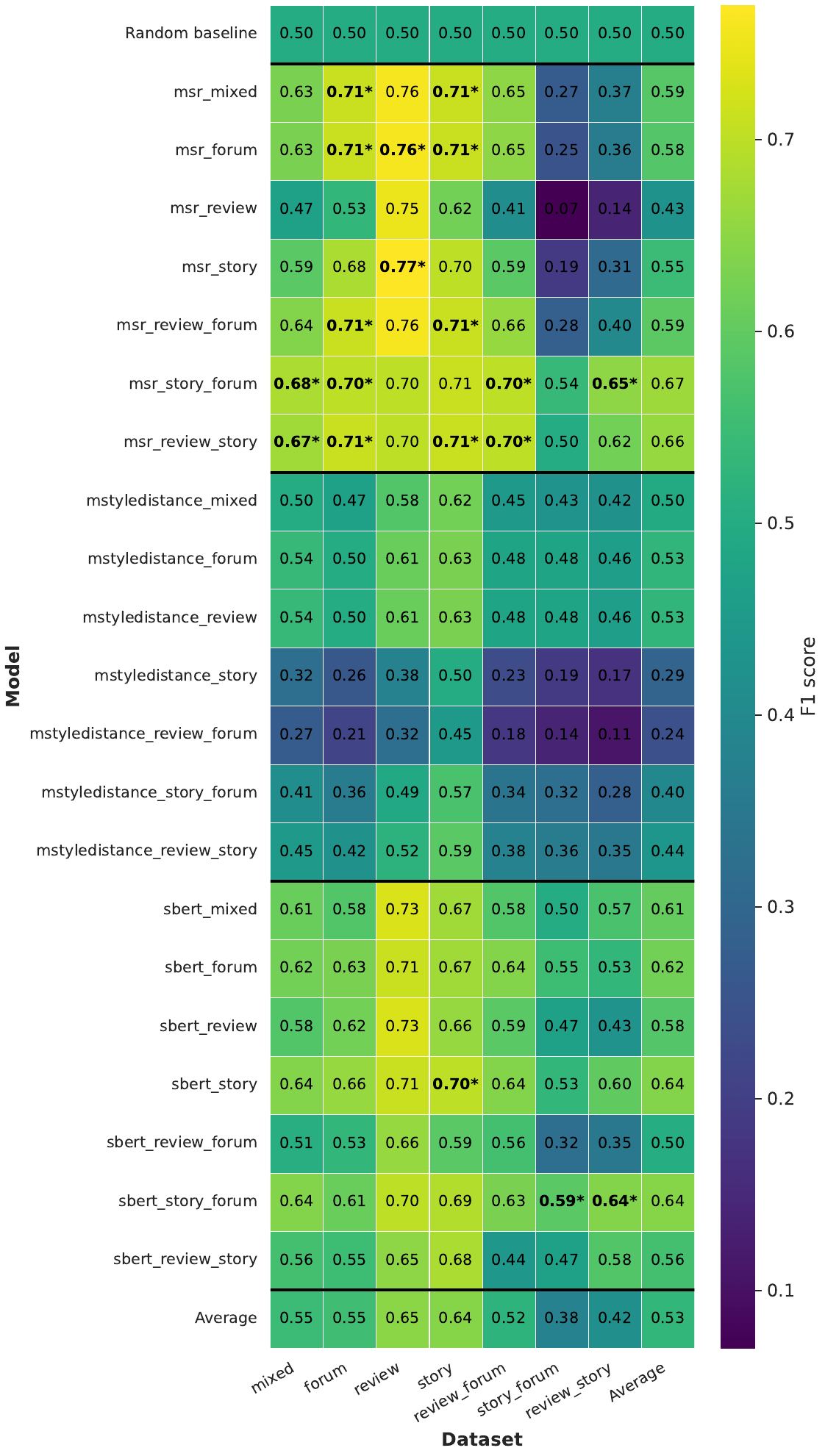}
        \caption{F1 scores}
        \label{fig:f1_all_names_embedding}
    \end{subfigure}%
    \hfill
    \begin{subfigure}[b]{0.48\textwidth}
        \centering
        \includegraphics[width=\linewidth]{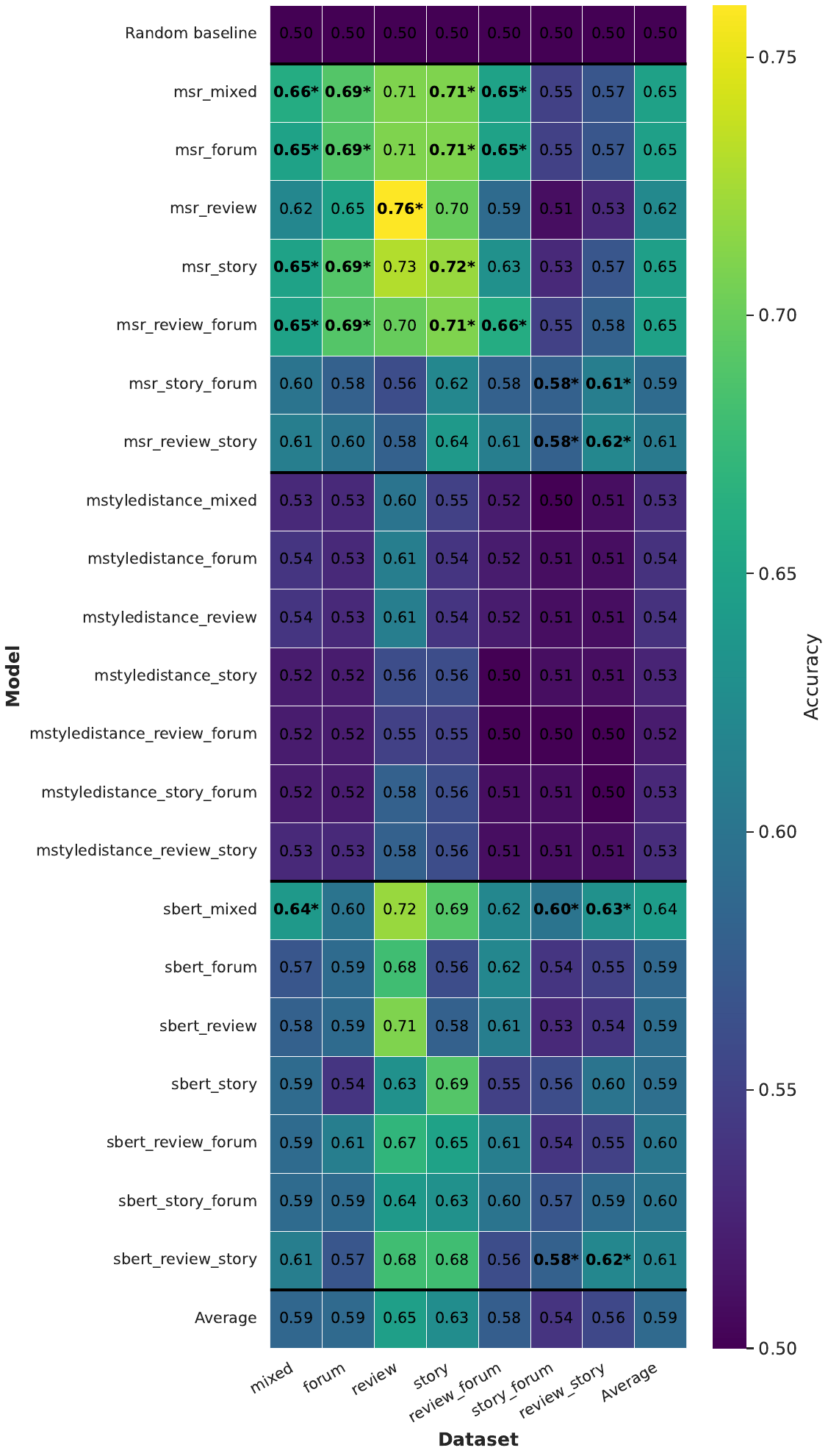}
        \caption{Accuracy scores}
        \label{fig:acc_all_names_embedding}
    \end{subfigure}

    \caption{Performance of all embedding-based models trained or calibrated and evaluated on all GenreShift train and test splits. Model names are shown on the y-axis and test dataset names on the x-axis. Each model name is followed by its training or calibration dataset. The best score for each test set is shown in bold. Statistically significant superiority over all other models is indicated by an asterisk (*; $p < 0.05$).}
    \label{fig:all_names_embedding}
\end{figure*}

\begin{figure}[h!]
    \centering
    \begin{subfigure}[b]{\linewidth}
        \centering
        \includegraphics[width=\linewidth]{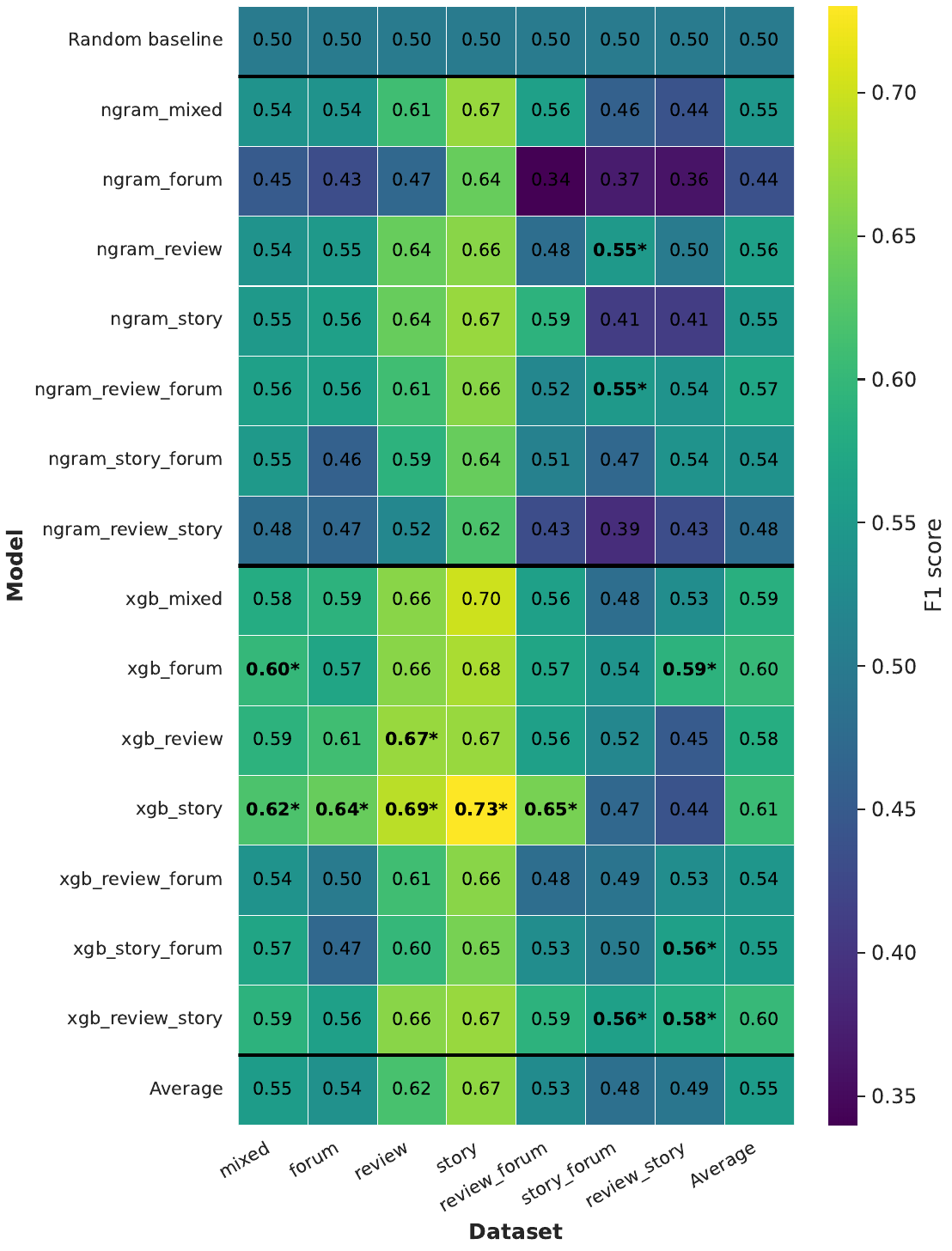}
        \caption{F1 scores}
        \label{fig:f1_all_names_feature}
    \end{subfigure}

    \vspace{0.5em}

    \begin{subfigure}[b]{\linewidth}
        \centering
        \includegraphics[width=\linewidth]{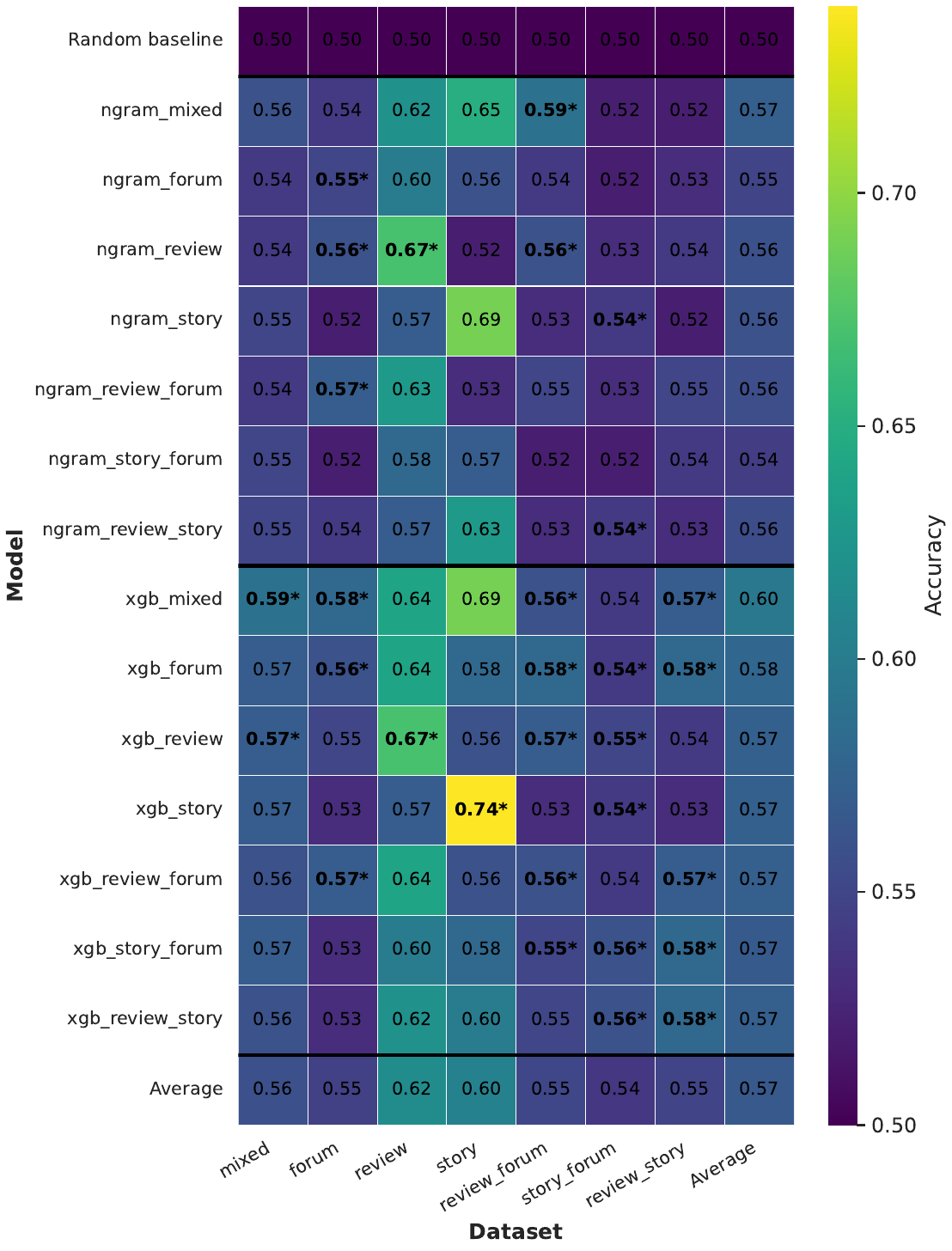}
        \caption{Accuracy scores}
        \label{fig:acc_all_names_feature}
    \end{subfigure}

    \caption{Performance of all feature-based models trained and evaluated on all GenreShift train and test splits. Model names are shown on the y-axis and test dataset names on the x-axis. Each model name is followed by its training dataset. Best scores are shown in bold. Statistically significant superiority over all other models is indicated by an asterisk (*; $p < 0.05$).}
    \label{fig:all_names_feature}
\end{figure}

\begin{figure}[h!]
    \centering
    \begin{subfigure}[b]{\linewidth}
        \centering
        \includegraphics[width=\linewidth]{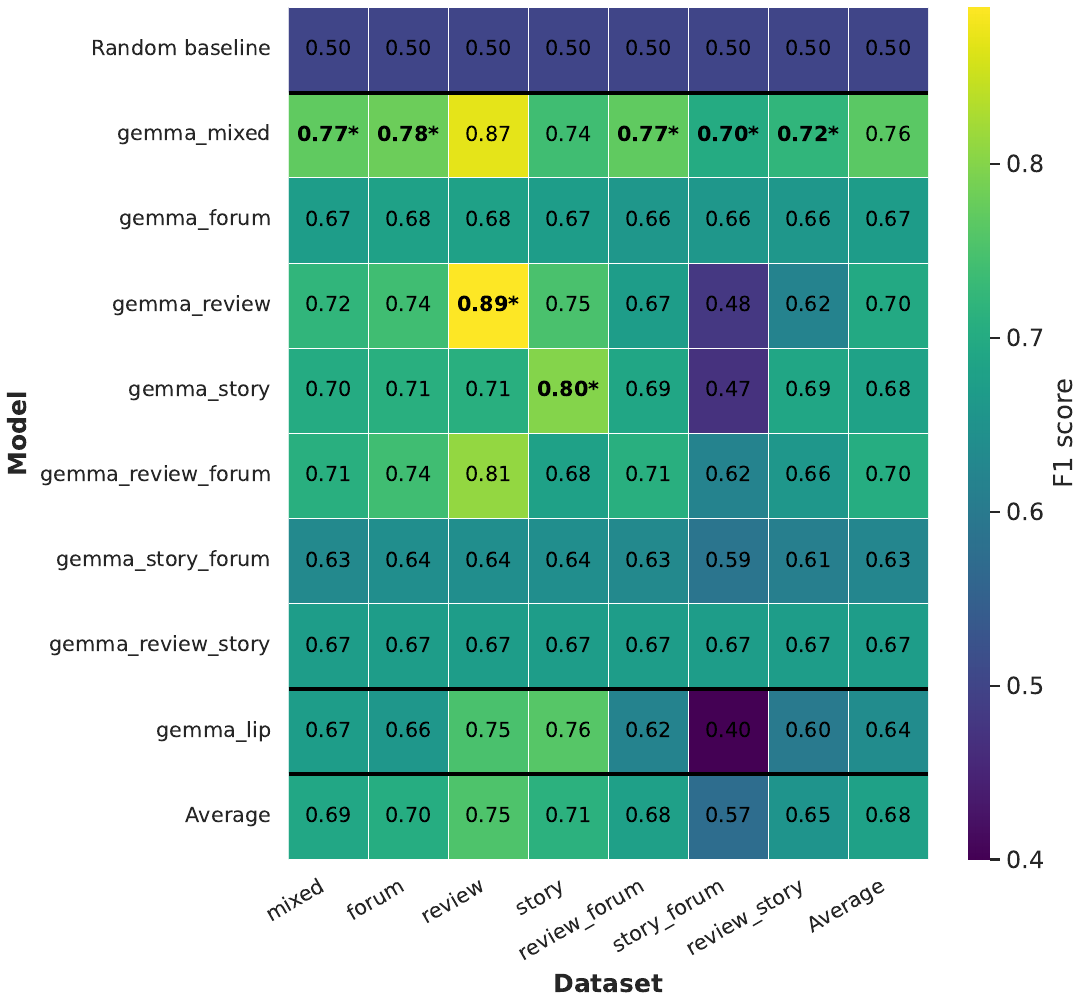}
        \caption{F1 scores}
        \label{fig:f1_all_names_llm}
    \end{subfigure}

    \vspace{0.5em}

    \begin{subfigure}[b]{\linewidth}
        \centering
        \includegraphics[width=\linewidth]{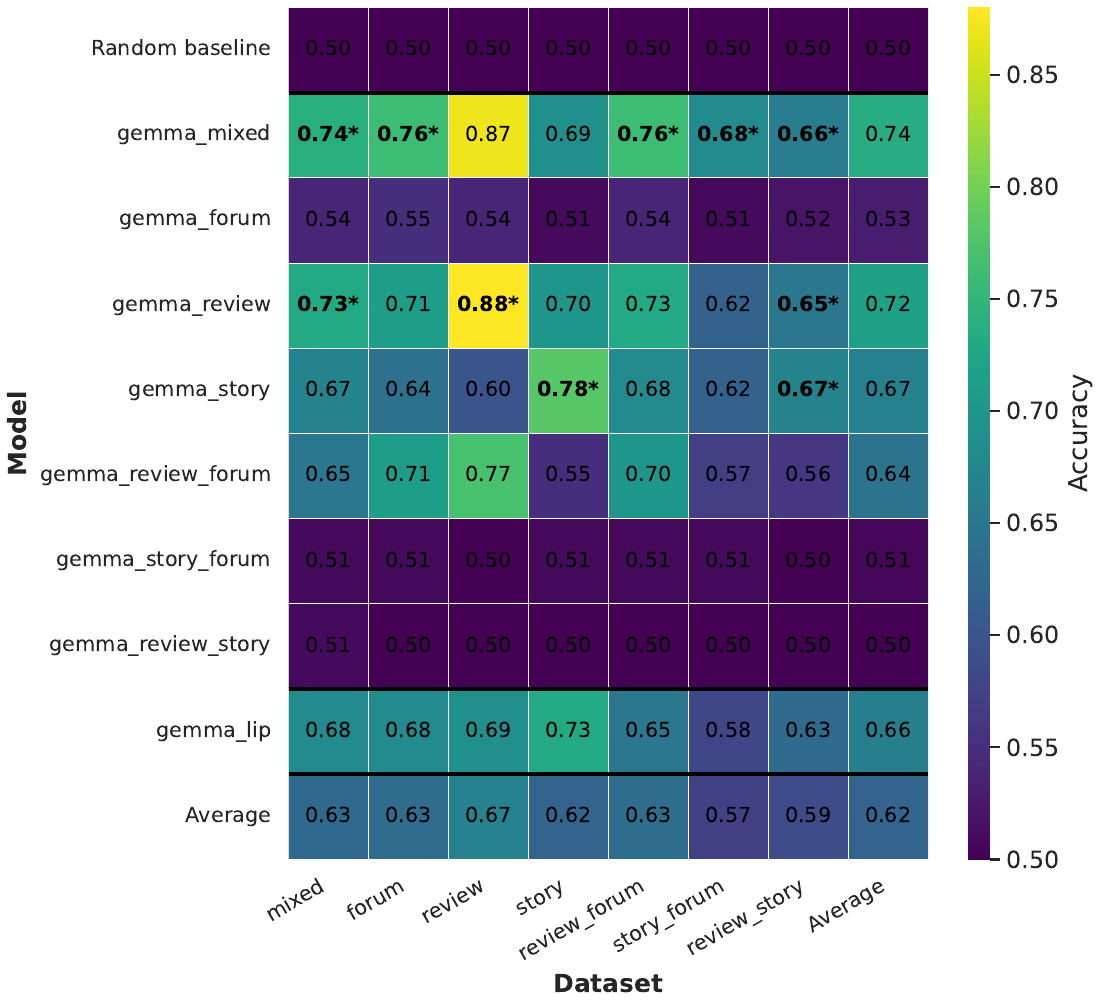}
        \caption{Accuracy scores}
        \label{fig:acc_all_names_llm}
    \end{subfigure}

    \caption{Performance of all LLM-based models trained and evaluated on all GenreShift train and test splits. Model names are shown on the y-axis and test dataset names on the x-axis. Each model name is followed by its training or calibration dataset. The best score for each test set is shown in bold. Statistically significant superiority over all other models is indicated by an asterisk (*; $p < 0.05$).}
    \label{fig:all_names_llm}
\end{figure}

We additionally compare F1 and accuracy scores between models within each methodological family to broaden our evaluation on AVShift. Figures \ref{fig:all_names_embedding}, \ref{fig:all_names_feature}, and \ref{fig:all_names_llm} show the full results for the embedding-based, feature-based, and LLM-based families, respectively, with F1 scores in (a) and accuracy scores in (b).

The results allow comparisons between individual models and their training or calibration configurations. In the LLM-based family (Figure \ref{fig:all_names_llm}), LoRA-tuned models consistently outperform LIP prompting across all evaluation datasets. Nevertheless, zero-shot LIP remains competitive across methodological families, outperforming all feature-based approaches and mStyleDistance on average (0.64 F1), demonstrating the potential of LLMs for AV even without task-specific fine-tuning.

In the feature-based family (Figure \ref{fig:all_names_feature}), XGB with its more extensive feature set outperforms the n-gram-based approach in F1 across all evaluation settings except Story-Forum, where both perform on par. Accuracy differences are less pronounced, with XGB significantly outperforming the n-gram approach in only three of seven settings.

Finally, in the embedding-based family (Figure \ref{fig:all_names_embedding}), both MSR and SBERT clearly outperform mStyleDistance. MSR significantly outperforms SBERT in F1 score in four of seven evaluation settings, performs on par in two, and is outperformed by SBERT on Story-Forum. In accuracy, MSR significantly outperforms SBERT in four settings and performs on par in the remaining three, making MSR the strongest embedding-based approach.

Based on their performance across most test datasets, we select the best-performing model from each family as its representative in the main paper.

\FloatBarrier

\FloatBarrier

\section{GenreShift Standardized Full Results}
\label{sec:full_results_genre_stand}
\begin{figure*}[t]
    \centering
    \begin{subfigure}[b]{0.48\textwidth}
        \centering
        \includegraphics[width=\linewidth]{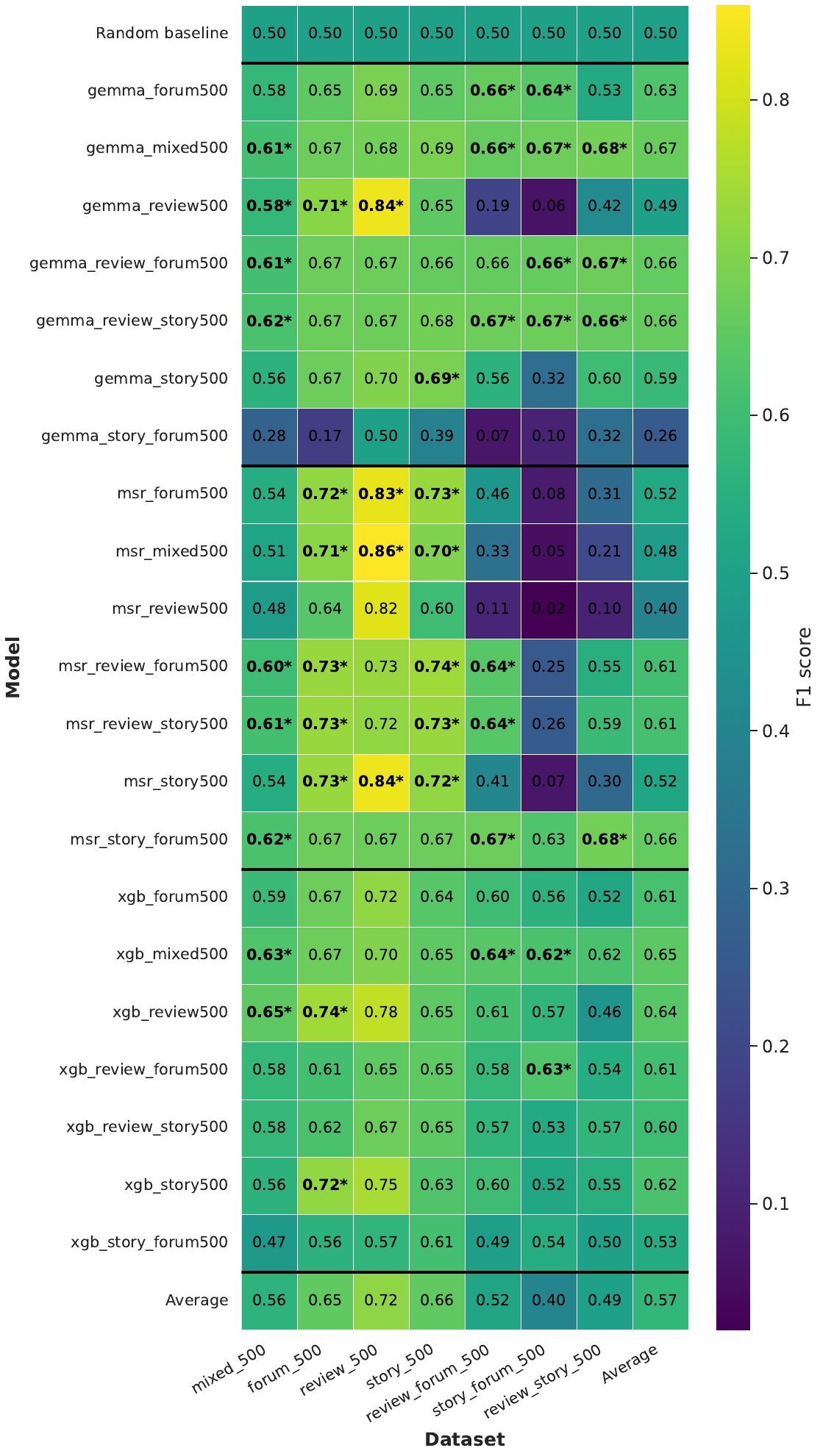}
        \caption{F1 scores}
        \label{fig:f1_500}
    \end{subfigure}%
    \hfill
    \begin{subfigure}[b]{0.48\textwidth}
        \centering
        \includegraphics[width=\linewidth]{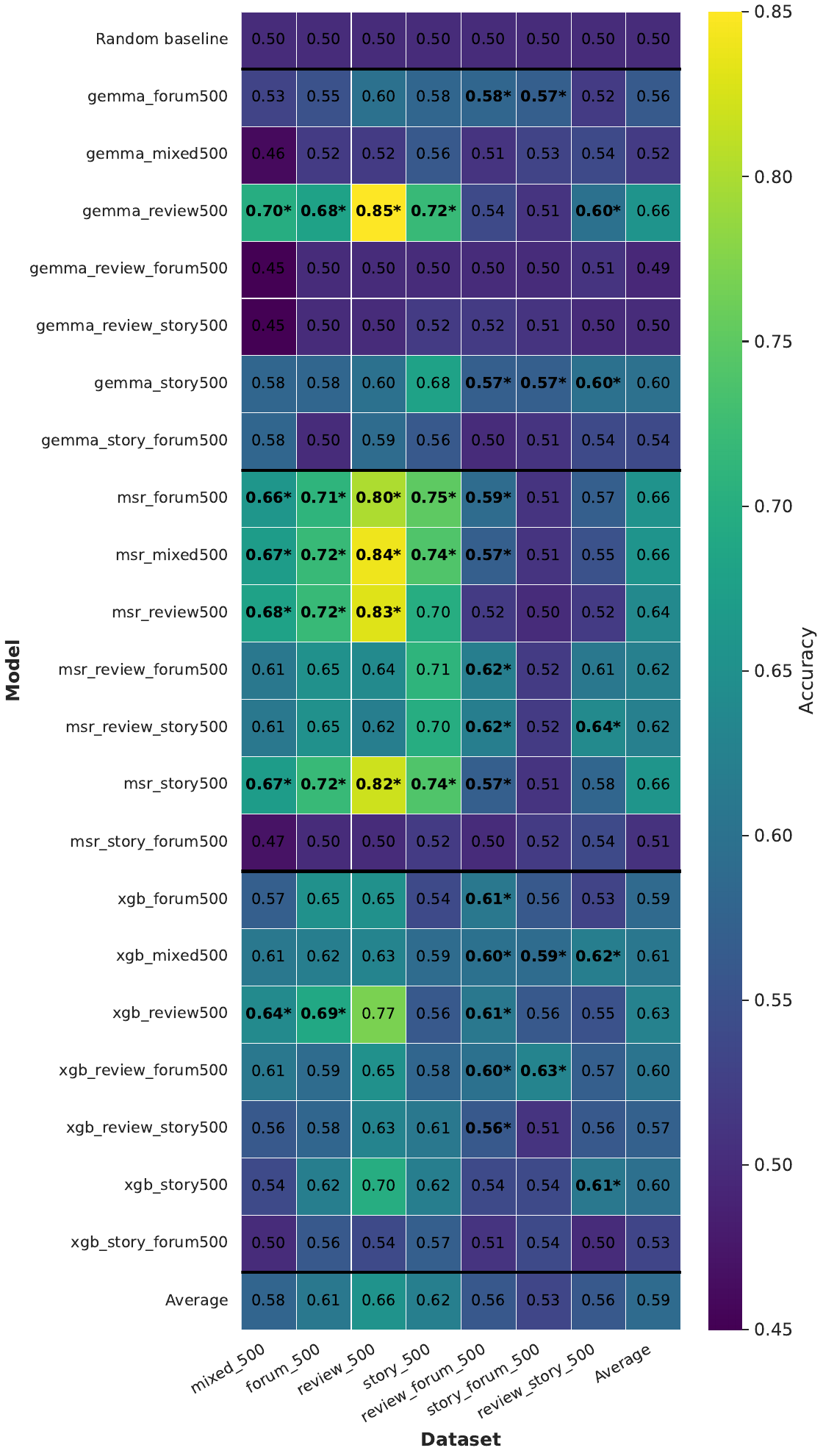}
        \caption{Accuracy scores}
        \label{fig:acc_500}
    \end{subfigure}

    \caption{Performance of the three representative models trained and evaluated on all standardized GenreShift train and test splits. Model names are shown on the y-axis and test dataset names on the x-axis. Each model name is followed by its training or calibration dataset. The best score for each test set is shown in bold. Statistically significant superiority over all other models is indicated by an asterisk (*; $p < 0.05$).}
    \label{fig:all_500_paper}
\end{figure*}

Figure \ref{fig:all_500_paper} presents the full F1 and accuracy results, respectively, on the standardized GenreShift benchmark. Compared to the unstandardized benchmark, Gemma no longer consistently outperforms the other approaches but shares the best performance with either MSR or XGB across all datasets. Performance in cross-genre settings drops significantly to a consistent F1 score of 0.67-0.68 across all genre pairs, indicating that larger training sizes are required for this more challenging setting. Furthermore, mixed training no longer provides a performance benefit in the standardized setting. However, the performance differences between genres in the in-domain setting remain consistent: review data again achieves substantially higher performance, despite document lengths and training sample sizes being standardized across all three genres.

\section{Most and Least Stable Features}
\label{sec:feature_ordering}
Table \ref{table:top_features} presents the top and bottom 30 features ranked by stability score across all three genre transfers (see Table \ref{table:xgb_features} for feature descriptions). Among the most stable features, we find several word bigrams consisting of function words, as well as part-of-speech (POS) trigrams, confirming earlier findings that function words and POS sequences represent stable indicators of writing style \citep{halvani_posnoise}. We further identify characteristic punctuation usage patterns, such as repeated exclamation marks.

Among the least stable features, we find, for example, average message length, reflecting the challenge of varying document lengths across different text genres. However, we also observe individual word bigrams, such as "als er", and POS trigrams, such as "punct-propn-verb", indicating that not all function word or POS sequences constitute stable style predictors. Furthermore, many character 4-grams appear among the least stable features, suggesting that character-level patterns may be more sensitive to genre-specific variations.

\begin{table}[h!]
\centering
\begin{adjustbox}{width=\linewidth}
\begin{tabular}{l|l}
\toprule
\makecell{\textbf{Top Features}}  &
\makecell{\textbf{Bottom Features}}  \\
\midrule
\texttt{mfw2\_aber in} & \texttt{mfc\_ kon} \\
\texttt{mfw2\_so als} & \texttt{mfc\_hien} \\
\texttt{nrPunct|} & \texttt{mfc\_sein} \\
\texttt{mfw2\_mal eine} & \texttt{mfc\_onnt} \\
\texttt{mfw2\_sein ich} & \texttt{mfc\_ bli} \\
\texttt{mfw2\_sollte man} & \texttt{mft\_punct propn verb} \\
\texttt{mfw2\_nicht als} & \texttt{mfw2\_als er} \\
\texttt{nrPunct\#} & \texttt{mfc\_Blic} \\
\texttt{mfw2\_ist doch} & \texttt{mfc\_ Er} \\
\texttt{mfw2\_von sich} & \texttt{mfc\_ sah} \\
\texttt{mfw2\_als das} & \texttt{mfc\_konn} \\
\texttt{mfw2\_als die} & \texttt{mfc\_sich} \\
\texttt{mfw2\_nur dass} & \texttt{mfc\_ er} \\
\texttt{mft\_sconj adj noun} & \texttt{mfc\_„Ich} \\
\texttt{nrPunct\textbackslash} & \texttt{mfc\_ „Ic} \\
\texttt{mfw2\_nach einem} & \texttt{mfc\_hiel} \\
\texttt{mfw2\_ist wie} & \texttt{wordLenDistri5} \\
\texttt{mft\_part punct aux} & \texttt{messageLen0} \\
\texttt{mfw2\_nicht gerade} & \texttt{mfc\_gte} \\
\texttt{mft\_sconj det det} & \texttt{mfc\_erte} \\
\texttt{mfw2\_zum beispiel} & \texttt{mft\_punct punct verb} \\
\texttt{mfw2\_mit einer} & \texttt{mfc\_egte} \\
\texttt{mfc\_!!!!} & \texttt{mfc\_elte} \\
\texttt{mfw2\_doch auch} & \texttt{mfc\_atte} \\
\texttt{mfw2\_nicht für} & \texttt{mfc\_te.} \\
\texttt{nrPunct+} & \texttt{mfc\_hatt} \\
\texttt{mfw2\_und alles} & \texttt{mfc\_kte} \\
\texttt{mfw2\_du sie} & \texttt{mfc\_ckte} \\
\texttt{mfc\_ Sel} & \texttt{mfc\_tete} \\
\texttt{mfw2\_wir sind} & \texttt{mfc\_ete} \\
\bottomrule
\end{tabular}
\end{adjustbox}
\caption{Top 30 most and least stable features across all genre transfers.}
\label{table:top_features}
\end{table}

\section{Pairwise Feature Analysis}
\label{sec:pairwise_feature_analysis}

Table \ref{table:pairwise_genre_stability_comp} reports the overlap of the 100 most and least stable features across different genre transitions. The overlap is generally low, ranging from 5\% to 27\% for the most stable features and from 2\% to 43\% for the least stable features. This indicates that no universally stable feature set exists across genre shifts, suggesting that feature stability should be analyzed separately for each genre pair, particularly when performing explicit feature selection. In contrast, Spearman rank correlations of feature stability remain moderate (0.42-0.72), indicating that while the most and least stable features vary substantially between genre pairs, the overall ranking of feature stability is comparatively consistent.

\begin{table}[h!]
\centering
\begin{adjustbox}{width=1\linewidth}
\begin{tabular}{ll|rrr} 
\toprule
 \makecell{\textbf{Pair1}} & \makecell{\textbf{Pair 2}} &\makecell{\textbf{Shared Top 100}} & \makecell{\textbf{Shared Bottom 100}} & \makecell{\textbf{Spearman Rank Correlation}} \\ 
\midrule 
Review-Forum & Story-Forum & 27\% & 3\% & 0.63 \\
Review-Forum & Review-Story & 8\% & 2\% & 0.42 \\ 
Story-Forum & Review-Story & 5\% & 43\% & 0.72 \\ 
\bottomrule
\end{tabular}
\end{adjustbox}
\caption{Differences in overlap of the 100 most and least stable features and relative stability feature ranking for each genre-pair combination.}
\label{table:pairwise_genre_stability_comp}
\end{table}

\section{Model and Data Licences}
\label{sec:licences}
We use all models in this work in accordance with their intended use as specified by their respective licences. Specifically, Gemma-4-31B-it is released under the Apache 2.0 licence \footnote{\url{https://ai.google.dev/gemma/apache_2}}, which permits modification, including fine-tuning, redistribution of fine-tuned models, and publication of research results.

To support reproducibility and future research on AV under distribution shifts while protecting authors' privacy, we provide access to the AVShift benchmark for research purposes through a gated-access process. Usernames are removed and replaced with pseudonymous identifiers to preserve author-level mappings. The data were collected and processed for non-commercial research purposes, relying on the text and data mining exception under \S~60d of the German Copyright Act (UrhG). They originate from publicly accessible pages that do not require user authentication. Access requests require applicants to provide information about themselves and their intended use of the data. Access is restricted to non-commercial research, prohibiting commercial use and redistribution. Please check our \href{https://github.com/NL2G/AVShift}{GitHub repository} for dataset access instructions. In addition, we release the complete preprocessing pipeline and scraping code to facilitate reproducibility and provide transparency regarding the dataset construction process.

%% file: custom.bib
@InProceedings{barlas_cross_domain,
author="Barlas, Georgios
and Stamatatos, Efstathios",
editor="Maglogiannis, Ilias
and Iliadis, Lazaros
and Pimenidis, Elias",
title="Cross-Domain Authorship Attribution Using Pre-trained Language Models",
booktitle="Artificial Intelligence Applications and Innovations",
year="2020",
publisher="Springer International Publishing",
address="Cham",
pages="255--266",
isbn="978-3-030-49161-1",
doi="https://doi.org/10.1007/978-3-030-49161-1_22"
}

@incollection{cafiero_too_old,
  title = {“I Am Too Old For This Style!” A Stylometric Benchmark of Age Effect
on Authorship Attribution},
  author = {Florian Cafiero and Lucence Ing and Simon Gabay and Thibault Clérice},
  year = {2025},
  booktitle = {Computational Humanities Research 2025},
  publisher = {Anthology of Computers and the Humanities},
  pages = {1248--1260},
  editor = {Taylor Arnold and Margherita Fantoli and Ruben Ros},
  doi = {10.63744/By09x5ZX3yWX}
}

@article{german-bert,
  title     = {{German BERT (bert-base-german-cased)}},
  author    = {Chan, Branden and M{\"o}ller, Timo and Pietsch, Malte and Soni, Tanay},
  year      = {2019},
  publisher = {Hugging Face},
  journal   = {Hugging Face repository},
  url       = {https://huggingface.co/google-bert/bert-base-german-cased},
  note      = {Accessed on 2025-12-31}
}

@inproceedings{halvani_posnoise,
author = {Halvani, Oren and Graner, Lukas},
title = {POSNoise: An Effective Countermeasure Against Topic Biases in Authorship Analysis},
year = {2021},
isbn = {9781450390514},
publisher = {Association for Computing Machinery},
address = {New York, NY, USA},
url = {https://doi.org/10.1145/3465481.3470050},
doi = {10.1145/3465481.3470050},
booktitle = {Proceedings of the 16th International Conference on Availability, Reliability and Security},
articleno = {47},
numpages = {12},
location = {Vienna, Austria},
series = {ARES '21}
}

@inproceedings{israeli-etal-2025-million,
    title = "The Million Authors Corpus: A Cross-Lingual and Cross-Domain {W}ikipedia Dataset for Authorship Verification",
    author = "Israeli, Abraham  and
      Liu, Shuai  and
      May, Jonathan  and
      Jurgens, David",
    editor = "Che, Wanxiang  and
      Nabende, Joyce  and
      Shutova, Ekaterina  and
      Pilehvar, Mohammad Taher",
    booktitle = "Findings of the Association for Computational Linguistics: ACL 2025",
    month = jul,
    year = "2025",
    address = "Vienna, Austria",
    publisher = "Association for Computational Linguistics",
    url = "https://aclanthology.org/2025.findings-acl.1335/",
    doi = "10.18653/v1/2025.findings-acl.1335",
    pages = "25997--26017",
    ISBN = "979-8-89176-256-5",
}

@inproceedings{kiefer-etal-2026-gerav,
    title = "{G}er{AV}: Towards New Heights in {G}erman Authorship Verification using Fine-Tuned {LLM}s on a New Benchmark",
    author = "Kiefer, Lotta  and
      Leiter, Christoph  and
      Takeshita, Sotaro  and
      Schmidt, Elena  and
      Eger, Steffen",
    editor = "Liakata, Maria  and
      Moreira, Viviane P.  and
      Zhang, Jiajun  and
      Jurgens, David",
    booktitle = "Findings of the {A}ssociation for {C}omputational {L}inguistics: {ACL} 2026",
    month = jul,
    year = "2026",
    address = "San Diego, California, United States",
    publisher = "Association for Computational Linguistics",
    url = "https://aclanthology.org/2026.findings-acl.1991/",
    doi = "10.18653/v1/2026.findings-acl.1991",
    pages = "40050--40069",
    ISBN = "979-8-89176-395-1",
}

@inproceedings{kim-etal-2025-msr,
    title = "Leveraging Multilingual Training for Authorship Representation: Enhancing Generalization across Languages and Domains",
    author = "Kim, Junghwan  and
      Zhang, Haotian  and
      Jurgens, David",
    editor = "Christodoulopoulos, Christos  and
      Chakraborty, Tanmoy  and
      Rose, Carolyn  and
      Peng, Violet",
    booktitle = "Proceedings of the 2025 Conference on Empirical Methods in Natural Language Processing",
    month = nov,
    year = "2025",
    address = "Suzhou, China",
    publisher = "Association for Computational Linguistics",
    url = "https://aclanthology.org/2025.emnlp-main.1766/",
    doi = "10.18653/v1/2025.emnlp-main.1766",
    pages = "34867--34892",
    ISBN = "979-8-89176-332-6",
}

@article{Ma_Le_Kang_Dou_Cadigan_Freitag_Ritter_Xu_2025, title={CROSSNEWS: A Cross-Genre Authorship Verification and Attribution Benchmark}, volume={39}, url={https://ojs.aaai.org/index.php/AAAI/article/view/34659}, DOI={10.1609/aaai.v39i23.34659}, number={23}, journal={Proceedings of the AAAI Conference on Artificial Intelligence}, author={Ma, Marcus and Le, Duong Minh and Kang, Junmo and Dou, Yao and Cadigan, John and Freitag, Dayne and Ritter, Alan and Xu, Wei}, year={2025}, month={Apr.}, pages={24777–24785} }

@article{overdorf_blogs_twitter,
author = {Overdorf, Rebekah and Greenstadt, Rachel},
year = {2016},
month = {07},
pages = {},
title = {Blogs, Twitter Feeds, and Reddit Comments: Cross-domain Authorship Attribution},
volume = {2016},
journal = {Proceedings on Privacy Enhancing Technologies},
doi = {10.1515/popets-2016-0021}
}

@inproceedings{richburg-etal-2024-ai-shift,
    title = "Automatic Authorship Analysis in Human-{AI} Collaborative Writing",
    author = "Richburg, Aquia  and
      Bao, Calvin  and
      Carpuat, Marine",
    editor = "Calzolari, Nicoletta  and
      Kan, Min-Yen  and
      Hoste, Veronique  and
      Lenci, Alessandro  and
      Sakti, Sakriani  and
      Xue, Nianwen",
    booktitle = "Proceedings of the 2024 Joint International Conference on Computational Linguistics, Language Resources and Evaluation (LREC-COLING 2024)",
    month = may,
    year = "2024",
    address = "Torino, Italia",
    publisher = "ELRA and ICCL",
    url = "https://aclanthology.org/2024.lrec-main.165/",
    pages = "1845--1855",
}

@article{stamatatos_masking_topic,
author = {Stamatatos, Efstathios},
title = {Masking topic-related information to enhance authorship attribution},
journal = {Journal of the Association for Information Science and Technology},
volume = {69},
number = {3},
pages = {461-473},
doi = {https://doi.org/10.1002/asi.23968},
url = {https://asistdl.onlinelibrary.wiley.com/doi/abs/10.1002/asi.23968},
eprint = {https://asistdl.onlinelibrary.wiley.com/doi/pdf/10.1002/asi.23968},
year = {2018}
}

@article{VANLEEUWEN2026100943,
title = {Cross-domain authorship verification with feature interaction networks: Evaluating no-holdout and holdout protocols},
journal = {Machine Learning with Applications},
volume = {25},
pages = {100943},
year = {2026},
issn = {2666-8270},
doi = {https://doi.org/10.1016/j.mlwa.2026.100943},
url = {https://www.sciencedirect.com/science/article/pii/S2666827026001088},
author = {Britt {van Leeuwen} and Sandjai Bhulai and Rob {van der Mei}},
}

@article{Yang_Zhu_Tang_Wang_2017_topic_drift, title={Authorship Attribution with Topic Drift Model}, volume={31}, url={https://ojs.aaai.org/index.php/AAAI/article/view/11062}, DOI={10.1609/aaai.v31i1.11062}, number={1}, journal={Proceedings of the AAAI Conference on Artificial Intelligence}, author={Yang, Min and Zhu, Dingju and Tang, Yong and Wang, Jingxuan}, year={2017}, month={Feb.} }

@InProceedings{stamatatos2022overview,
  author =                   {Efstathios Stamatatos and Mike Kestemont and Krzysztof Kredens and Piotr Pezik and Annina Heini and Janek Bevendorff and Benno Stein and Martin Potthast},
  booktitle =                {{CLEF} 2022 Labs and Workshops, Notebook Papers},
  editor =                   {Guglielmo Faggioli and Nicola Ferro and Allan Hanbury and Martin Potthast},
  ids =                      {potthast:2022l},
  issn =                     {1613-0073},
  month =                    sep,
  numpages =                 13,
  pages =                    {2301-2313},
  publisher =                {CEUR-WS.org},
  series =                   {CEUR Workshop Proceedings},
  site =                     {Bologna, Italy},
  title =                    {{Overview of the Authorship Verification Task at PAN 2022}},
  url =                      {https://ceur-ws.org/Vol-3180/paper-184.pdf},
  volume =                   3180,
  year =                     2022
}

@inproceedings{Stamatatos2023OverviewOT,
  author    = {Efstathios Stamatatos and
               Krzysztof Kredens and
               Piotr Pezik and
               Annina Heini and
               Janek Bevendorff and
               Benno Stein and
               Martin Potthast},
  editor    = {Mohammad Aliannejadi and
               Guglielmo Faggioli and
               Nicola Ferro and
               Michail Vlachos},
  title     = {{Overview of the Authorship Verification Task at PAN 2023}},
  booktitle = {{Working Notes of the Conference and Labs of the Evaluation Forum (CLEF 2023), Thessaloniki, Greece, September 18th to 21st, 2023}},
  series    = {CEUR Workshop Proceedings},
  volume    = {3497},
  pages     = {2476--2491},
  publisher = {CEUR-WS.org},
  year      = {2023},
  url       = {https://ceur-ws.org/Vol-3497/paper-199.pdf}
}

@article{hu2023tdrlm5,
title = {TDRLM: Stylometric learning for authorship verification by Topic-Debiasing},
journal = {Expert Systems with Applications},
volume = {233},
pages = {120745},
year = {2023},
issn = {0957-4174},
doi = {https://doi.org/10.1016/j.eswa.2023.120745},
url = {https://www.sciencedirect.com/science/article/pii/S0957417423012472},
author = {Xinyu Hu and Weihan Ou and Sudipta Acharya and Steven H.H. Ding and Ryan D’Gama and Hanbo Yu},
}

@inproceedings{Zhang2021ImprovingAV,
  title={Improving Authorship Verification using Linguistic Divergence},
  author={Yifan Zhang and Dainis Boumber and Marjan Hosseinia and Fan Yang and Arjun Mukherjee},
  booktitle={ROMCIR@ECIR},
  year={2021},
  url={https://api.semanticscholar.org/CorpusID:232223310}
}

@inproceedings{azarbonyad2015time,
author = {Azarbonyad, Hosein and Dehghani, Mostafa and Marx, Maarten and Kamps, Jaap},
title = {Time-Aware Authorship Attribution for Short Text Streams},
year = {2015},
isbn = {9781450336215},
publisher = {Association for Computing Machinery},
address = {New York, NY, USA},
url = {https://doi.org/10.1145/2766462.2767799},
doi = {10.1145/2766462.2767799},
pages = {727–730},
numpages = {4},
location = {Santiago, Chile},
series = {SIGIR '15}
}

@inproceedings{mysore-etal-2025-prototypical,
    title = "Prototypical Human-{AI} Collaboration Behaviors from {LLM}-Assisted Writing in the Wild",
    author = "Mysore, Sheshera  and
      Das, Debarati  and
      Cao, Hancheng  and
      Sarrafzadeh, Bahareh",
    booktitle = "Proceedings of the 2025 Conference on Empirical Methods in Natural Language Processing",
    month = nov,
    year = "2025",
    address = "Suzhou, China",
    publisher = "Association for Computational Linguistics",
    url = "https://aclanthology.org/2025.emnlp-main.852/",
    doi = "10.18653/v1/2025.emnlp-main.852",
    pages = "16819--16846",
    ISBN = "979-8-89176-332-6"
}

@inproceedings{lee2022coauthor, series={CHI ’22},
   title={CoAuthor: Designing a Human-AI Collaborative Writing Dataset for Exploring Language Model Capabilities},
   url={http://dx.doi.org/10.1145/3491102.3502030},
   DOI={10.1145/3491102.3502030},
   booktitle={CHI Conference on Human Factors in Computing Systems},
   publisher={ACM},
   author={Lee, Mina and Liang, Percy and Yang, Qian},
   year={2022},
   month=Apr, pages={1–19},
   collection={CHI ’22} }

@article{coulthard_idiolect_2004,
    author = {Coulthard, Malcolm},
    title = {Author Identification, Idiolect, and Linguistic Uniqueness},
    journal = {Applied Linguistics},
    volume = {25},
    number = {4},
    pages = {431-447},
    year = {2004},
    month = {12},
    issn = {0142-6001},
    doi = {10.1093/applin/25.4.431},
    url = {https://doi.org/10.1093/applin/25.4.431},
    eprint = {https://academic.oup.com/applij/article-pdf/25/4/431/480988/250431.pdf},
}

@book{Nini_2023, place={Cambridge}, series={Elements in Forensic Linguistics}, title={A Theory of Linguistic Individuality for Authorship Analysis}, 
doi={https://doi.org/10.1017/9781108974851 },
publisher={Cambridge University Press}, 
author={Nini, Andrea}, 
year={2023}, 
collection={Elements in Forensic Linguistics}}

@inproceedings{kestemont2020overview,
  author    = {Mike Kestemont and
               Enrique Manjavacas and
               Ilia Markov and
               Janek Bevendorff and
               Matti Wiegmann and
               Efstathios Stamatatos and
               Martin Potthast and
               Benno Stein},
  editor    = {Linda Cappellato and
               Carsten Eickhoff and
               Nicola Ferro and
               Aur\'{e}lie N\'{e}v\'{e}ol},
  title     = {{Overview of the Cross-Domain Authorship Verification Task at PAN 2020}},
  booktitle = {{Working Notes of CLEF 2020 - Conference and Labs of the Evaluation Forum, Thessaloniki, Greece, September 22-25, 2020}},
  series    = {CEUR Workshop Proceedings},
  volume    = {2696},
  publisher = {CEUR-WS.org},
  year      = {2020},
  url       = {https://ceur-ws.org/Vol-2696/paper_264.pdf}
}

@article{lewis_news_benchmark,
  title={RCV1: A New Benchmark Collection for Text Categorization Research},
  author={David D. Lewis and Yiming Yang and Tony G. Rose and Fan Li},
  journal={J. Mach. Learn. Res.},
  year={2004},
  volume={5},
  pages={361-397},
  url={https://api.semanticscholar.org/CorpusID:11027141}
}

@article{sourati_shrinking_2026,
	title = {The shrinking landscape of linguistic diversity in the age of large language models},
	copyright = {2026 The Author(s), under exclusive licence to Springer Nature Limited},
	issn = {2397-3374},
	url = {https://www.nature.com/articles/s41562-026-02550-0},
	doi = {10.1038/s41562-026-02550-0},
	language = {en},
	urldate = {2026-09-13},
	journal = {Nature Human Behaviour},
	publisher = {Nature Publishing Group},
	author = {Sourati, Zhivar and Karimi-Malekabadi, Farzan and Ozcan, Meltem and McDaniel, Colin and Ziabari, Alireza and Trager, Jackson and Tak, Ala N. and Chen, Meng and Morstatter, Fred and Dehghani, Morteza},
	month = aug,
	year = {2026},
	pages = {1--18},
}

@inproceedings{manolache_veridark,
 author = {Manolache, Andrei and Brad, Florin and Barbalau, Antonio and Ionescu, Radu Tudor and Popescu, Marius},
 booktitle = {Advances in Neural Information Processing Systems},
 doi = {10.52202/068431-1133},
 editor = {S. Koyejo and S. Mohamed and A. Agarwal and D. Belgrave and K. Cho and A. Oh},
 pages = {15574--15588},
 publisher = {Curran Associates, Inc.},
 title = {VeriDark: A Large-Scale Benchmark for Authorship Verification on the Dark Web},
 url = {https://proceedings.neurips.cc/paper_files/paper/2022/file/64008fa30cba9b4d1ab1bd3bd3d57d61-Paper-Datasets_and_Benchmarks.pdf},
 volume = {35},
 year = {2022}
}

@article{eder2011stylefingerprints,
  author  = {Eder, Maciej},
  title   = {Style-Markers in Authorship Attribution: A Cross-Language Study of the Authorial Fingerprint},
  journal = {Studies in Polish Linguistics},
  year    = {2011},
  volume  = {6},
  number  = {1},
  pages   = {99--114},
  url     = {https://ejournals.eu/en/journal_article_files/full_text/018ecec1-46ce-719e-af43-df8ff955dba5/download}
}

@book{kreuz2023linguistic,
  title={Linguistic fingerprints: How language creates and reveals identity},
  author={Kreuz, Roger},
  year={2023},
  publisher={Simon and Schuster},
  isbn = {978-1-63388-897-5}
}

@software{beautifulsoup,
  author = {Richardson, Leonard},
  title = {Beautiful Soup Documentation},
  url = {https://beautiful-soup.readthedocs.io/en/latest/},
  version = {4.14.3},
  year = {2025}
}

@misc{grattafiori2024llama3herdmodels,
      title={The Llama 3 Herd of Models}, 
      author={Aaron Grattafiori and Abhimanyu Dubey and Abhinav Jauhri and Abhinav Pandey and Abhishek Kadian and Ahmad Al-Dahle and Aiesha Letman and Akhil Mathur and Alan Schelten and Alex Vaughan and Amy Yang and Angela Fan and Anirudh Goyal and Anthony Hartshorn and Aobo Yang and Archi Mitra and Archie Sravankumar and Artem Korenev and Arthur Hinsvark and Arun Rao and Aston Zhang and Aurelien Rodriguez and Austen Gregerson and Ava Spataru and Baptiste Roziere and Bethany Biron and Binh Tang and Bobbie Chern and Charlotte Caucheteux and Chaya Nayak and Chloe Bi and Chris Marra and Chris McConnell and Christian Keller and Christophe Touret and Chunyang Wu and Corinne Wong and Cristian Canton Ferrer and Cyrus Nikolaidis and Damien Allonsius and Daniel Song and Danielle Pintz and Danny Livshits and Danny Wyatt and David Esiobu and Dhruv Choudhary and Dhruv Mahajan and Diego Garcia-Olano and Diego Perino and Dieuwke Hupkes and Egor Lakomkin and Ehab AlBadawy and Elina Lobanova and Emily Dinan and Eric Michael Smith and Filip Radenovic and Francisco Guzmán and Frank Zhang and Gabriel Synnaeve and Gabrielle Lee and Georgia Lewis Anderson and Govind Thattai and Graeme Nail and Gregoire Mialon and Guan Pang and Guillem Cucurell and Hailey Nguyen and Hannah Korevaar and Hu Xu and Hugo Touvron and Iliyan Zarov and Imanol Arrieta Ibarra and Isabel Kloumann and Ishan Misra and Ivan Evtimov and Jack Zhang and Jade Copet and Jaewon Lee and Jan Geffert and Jana Vranes and Jason Park and Jay Mahadeokar and Jeet Shah and Jelmer van der Linde and Jennifer Billock and Jenny Hong and Jenya Lee and Jeremy Fu and Jianfeng Chi and Jianyu Huang and Jiawen Liu and Jie Wang and Jiecao Yu and Joanna Bitton and Joe Spisak and Jongsoo Park and Joseph Rocca and Joshua Johnstun and Joshua Saxe and Junteng Jia and Kalyan Vasuden Alwala and Karthik Prasad and Kartikeya Upasani and Kate Plawiak and Ke Li and Kenneth Heafield and Kevin Stone and Khalid El-Arini and Krithika Iyer and Kshitiz Malik and Kuenley Chiu and Kunal Bhalla and Kushal Lakhotia and Lauren Rantala-Yeary and Laurens van der Maaten and Lawrence Chen and Liang Tan and Liz Jenkins and Louis Martin and Lovish Madaan and Lubo Malo and Lukas Blecher and Lukas Landzaat and Luke de Oliveira and Madeline Muzzi and Mahesh Pasupuleti and Mannat Singh and Manohar Paluri and Marcin Kardas and Maria Tsimpoukelli and Mathew Oldham and Mathieu Rita and Maya Pavlova and Melanie Kambadur and Mike Lewis and Min Si and Mitesh Kumar Singh and Mona Hassan and Naman Goyal and Narjes Torabi and Nikolay Bashlykov and Nikolay Bogoychev and Niladri Chatterji and Ning Zhang and Olivier Duchenne and Onur Çelebi and Patrick Alrassy and Pengchuan Zhang and Pengwei Li and Petar Vasic and Peter Weng and Prajjwal Bhargava and Pratik Dubal and Praveen Krishnan and Punit Singh Koura and Puxin Xu and Qing He and Qingxiao Dong and Ragavan Srinivasan and Raj Ganapathy and Ramon Calderer and Ricardo Silveira Cabral and Robert Stojnic and Roberta Raileanu and Rohan Maheswari and Rohit Girdhar and Rohit Patel and Romain Sauvestre and Ronnie Polidoro and Roshan Sumbaly and Ross Taylor and Ruan Silva and Rui Hou and Rui Wang and Saghar Hosseini and Sahana Chennabasappa and Sanjay Singh and Sean Bell and Seohyun Sonia Kim and Sergey Edunov and Shaoliang Nie and Sharan Narang and Sharath Raparthy and Sheng Shen and Shengye Wan and Shruti Bhosale and Shun Zhang and Simon Vandenhende and Soumya Batra and Spencer Whitman and Sten Sootla and Stephane Collot and Suchin Gururangan and Sydney Borodinsky and Tamar Herman and Tara Fowler and Tarek Sheasha and Thomas Georgiou and Thomas Scialom and Tobias Speckbacher and Todor Mihaylov and Tong Xiao and Ujjwal Karn and Vedanuj Goswami and Vibhor Gupta and Vignesh Ramanathan and Viktor Kerkez and Vincent Gonguet and Virginie Do and Vish Vogeti and Vítor Albiero and Vladan Petrovic and Weiwei Chu and Wenhan Xiong and Wenyin Fu and Whitney Meers and Xavier Martinet and Xiaodong Wang and Xiaofang Wang and Xiaoqing Ellen Tan and Xide Xia and Xinfeng Xie and Xuchao Jia and Xuewei Wang and Yaelle Goldschlag and Yashesh Gaur and Yasmine Babaei and Yi Wen and Yiwen Song and Yuchen Zhang and Yue Li and Yuning Mao and Zacharie Delpierre Coudert and Zheng Yan and Zhengxing Chen and Zoe Papakipos and Aaditya Singh and Aayushi Srivastava and Abha Jain and Adam Kelsey and Adam Shajnfeld and Adithya Gangidi and Adolfo Victoria and Ahuva Goldstand and Ajay Menon and Ajay Sharma and Alex Boesenberg and Alexei Baevski and Allie Feinstein and Amanda Kallet and Amit Sangani and Amos Teo and Anam Yunus and Andrei Lupu and Andres Alvarado and Andrew Caples and Andrew Gu and Andrew Ho and Andrew Poulton and Andrew Ryan and Ankit Ramchandani and Annie Dong and Annie Franco and Anuj Goyal and Aparajita Saraf and Arkabandhu Chowdhury and Ashley Gabriel and Ashwin Bharambe and Assaf Eisenman and Azadeh Yazdan and Beau James and Ben Maurer and Benjamin Leonhardi and Bernie Huang and Beth Loyd and Beto De Paola and Bhargavi Paranjape and Bing Liu and Bo Wu and Boyu Ni and Braden Hancock and Bram Wasti and Brandon Spence and Brani Stojkovic and Brian Gamido and Britt Montalvo and Carl Parker and Carly Burton and Catalina Mejia and Ce Liu and Changhan Wang and Changkyu Kim and Chao Zhou and Chester Hu and Ching-Hsiang Chu and Chris Cai and Chris Tindal and Christoph Feichtenhofer and Cynthia Gao and Damon Civin and Dana Beaty and Daniel Kreymer and Daniel Li and David Adkins and David Xu and Davide Testuggine and Delia David and Devi Parikh and Diana Liskovich and Didem Foss and Dingkang Wang and Duc Le and Dustin Holland and Edward Dowling and Eissa Jamil and Elaine Montgomery and Eleonora Presani and Emily Hahn and Emily Wood and Eric-Tuan Le and Erik Brinkman and Esteban Arcaute and Evan Dunbar and Evan Smothers and Fei Sun and Felix Kreuk and Feng Tian and Filippos Kokkinos and Firat Ozgenel and Francesco Caggioni and Frank Kanayet and Frank Seide and Gabriela Medina Florez and Gabriella Schwarz and Gada Badeer and Georgia Swee and Gil Halpern and Grant Herman and Grigory Sizov and Guangyi and Zhang and Guna Lakshminarayanan and Hakan Inan and Hamid Shojanazeri and Han Zou and Hannah Wang and Hanwen Zha and Haroun Habeeb and Harrison Rudolph and Helen Suk and Henry Aspegren and Hunter Goldman and Hongyuan Zhan and Ibrahim Damlaj and Igor Molybog and Igor Tufanov and Ilias Leontiadis and Irina-Elena Veliche and Itai Gat and Jake Weissman and James Geboski and James Kohli and Janice Lam and Japhet Asher and Jean-Baptiste Gaya and Jeff Marcus and Jeff Tang and Jennifer Chan and Jenny Zhen and Jeremy Reizenstein and Jeremy Teboul and Jessica Zhong and Jian Jin and Jingyi Yang and Joe Cummings and Jon Carvill and Jon Shepard and Jonathan McPhie and Jonathan Torres and Josh Ginsburg and Junjie Wang and Kai Wu and Kam Hou U and Karan Saxena and Kartikay Khandelwal and Katayoun Zand and Kathy Matosich and Kaushik Veeraraghavan and Kelly Michelena and Keqian Li and Kiran Jagadeesh and Kun Huang and Kunal Chawla and Kyle Huang and Lailin Chen and Lakshya Garg and Lavender A and Leandro Silva and Lee Bell and Lei Zhang and Liangpeng Guo and Licheng Yu and Liron Moshkovich and Luca Wehrstedt and Madian Khabsa and Manav Avalani and Manish Bhatt and Martynas Mankus and Matan Hasson and Matthew Lennie and Matthias Reso and Maxim Groshev and Maxim Naumov and Maya Lathi and Meghan Keneally and Miao Liu and Michael L. Seltzer and Michal Valko and Michelle Restrepo and Mihir Patel and Mik Vyatskov and Mikayel Samvelyan and Mike Clark and Mike Macey and Mike Wang and Miquel Jubert Hermoso and Mo Metanat and Mohammad Rastegari and Munish Bansal and Nandhini Santhanam and Natascha Parks and Natasha White and Navyata Bawa and Nayan Singhal and Nick Egebo and Nicolas Usunier and Nikhil Mehta and Nikolay Pavlovich Laptev and Ning Dong and Norman Cheng and Oleg Chernoguz and Olivia Hart and Omkar Salpekar and Ozlem Kalinli and Parkin Kent and Parth Parekh and Paul Saab and Pavan Balaji and Pedro Rittner and Philip Bontrager and Pierre Roux and Piotr Dollar and Polina Zvyagina and Prashant Ratanchandani and Pritish Yuvraj and Qian Liang and Rachad Alao and Rachel Rodriguez and Rafi Ayub and Raghotham Murthy and Raghu Nayani and Rahul Mitra and Rangaprabhu Parthasarathy and Raymond Li and Rebekkah Hogan and Robin Battey and Rocky Wang and Russ Howes and Ruty Rinott and Sachin Mehta and Sachin Siby and Sai Jayesh Bondu and Samyak Datta and Sara Chugh and Sara Hunt and Sargun Dhillon and Sasha Sidorov and Satadru Pan and Saurabh Mahajan and Saurabh Verma and Seiji Yamamoto and Sharadh Ramaswamy and Shaun Lindsay and Shaun Lindsay and Sheng Feng and Shenghao Lin and Shengxin Cindy Zha and Shishir Patil and Shiva Shankar and Shuqiang Zhang and Shuqiang Zhang and Sinong Wang and Sneha Agarwal and Soji Sajuyigbe and Soumith Chintala and Stephanie Max and Stephen Chen and Steve Kehoe and Steve Satterfield and Sudarshan Govindaprasad and Sumit Gupta and Summer Deng and Sungmin Cho and Sunny Virk and Suraj Subramanian and Sy Choudhury and Sydney Goldman and Tal Remez and Tamar Glaser and Tamara Best and Thilo Koehler and Thomas Robinson and Tianhe Li and Tianjun Zhang and Tim Matthews and Timothy Chou and Tzook Shaked and Varun Vontimitta and Victoria Ajayi and Victoria Montanez and Vijai Mohan and Vinay Satish Kumar and Vishal Mangla and Vlad Ionescu and Vlad Poenaru and Vlad Tiberiu Mihailescu and Vladimir Ivanov and Wei Li and Wenchen Wang and Wenwen Jiang and Wes Bouaziz and Will Constable and Xiaocheng Tang and Xiaojian Wu and Xiaolan Wang and Xilun Wu and Xinbo Gao and Yaniv Kleinman and Yanjun Chen and Ye Hu and Ye Jia and Ye Qi and Yenda Li and Yilin Zhang and Ying Zhang and Yossi Adi and Youngjin Nam and Yu and Wang and Yu Zhao and Yuchen Hao and Yundi Qian and Yunlu Li and Yuzi He and Zach Rait and Zachary DeVito and Zef Rosnbrick and Zhaoduo Wen and Zhenyu Yang and Zhiwei Zhao and Zhiyu Ma},
      year={2024},
      eprint={2407.21783},
      archivePrefix={arXiv},
      primaryClass={cs.AI},
      url={https://arxiv.org/abs/2407.21783}, 
}

@inproceedings{wang-etal-2024-improving-text-mistral,
    title = "Improving Text Embeddings with Large Language Models",
    author = "Wang, Liang  and
      Yang, Nan  and
      Huang, Xiaolong  and
      Yang, Linjun  and
      Majumder, Rangan  and
      Wei, Furu",
    editor = "Ku, Lun-Wei  and
      Martins, Andre  and
      Srikumar, Vivek",
    booktitle = "Proceedings of the 62nd Annual Meeting of the Association for Computational Linguistics (Volume 1: Long Papers)",
    month = aug,
    year = "2024",
    address = "Bangkok, Thailand",
    publisher = "Association for Computational Linguistics",
    url = "https://aclanthology.org/2024.acl-long.642/",
    doi = "10.18653/v1/2024.acl-long.642",
    pages = "11897--11916",
}

@inproceedings{weerasinghe2021feature,
  title={Feature Vector Difference based Authorship Verification for Open-World Settings.},
  author={Weerasinghe, Janith and Singh, Rhia and Greenstadt, Rachel},
  booktitle={CLEF (Working Notes)},
  pages={2201--2207},
  year={2021},
  url={https://ceur-ws.org/Vol-2936/paper-197.pdf}
}

@misc{gemmateam2025gemma3technicalreport,
      title={Gemma 3 Technical Report}, 
      author={{Gemma Team} and Aishwarya Kamath and Johan Ferret and Shreya Pathak and Nino Vieillard and Ramona Merhej and Sarah Perrin and Tatiana Matejovicova and Alexandre Ramé and Morgane Rivière and Louis Rouillard and Thomas Mesnard and Geoffrey Cideron and Jean-bastien Grill and Sabela Ramos and Edouard Yvinec and Michelle Casbon and Etienne Pot and Ivo Penchev and Gaël Liu and Francesco Visin and Kathleen Kenealy and Lucas Beyer and Xiaohai Zhai and Anton Tsitsulin and Robert Busa-Fekete and Alex Feng and Noveen Sachdeva and Benjamin Coleman and Yi Gao and Basil Mustafa and Iain Barr and Emilio Parisotto and David Tian and Matan Eyal and Colin Cherry and Jan-Thorsten Peter and Danila Sinopalnikov and Surya Bhupatiraju and Rishabh Agarwal and Mehran Kazemi and Dan Malkin and Ravin Kumar and David Vilar and Idan Brusilovsky and Jiaming Luo and Andreas Steiner and Abe Friesen and Abhanshu Sharma and Abheesht Sharma and Adi Mayrav Gilady and Adrian Goedeckemeyer and Alaa Saade and Alex Feng and Alexander Kolesnikov and Alexei Bendebury and Alvin Abdagic and Amit Vadi and András György and André Susano Pinto and Anil Das and Ankur Bapna and Antoine Miech and Antoine Yang and Antonia Paterson and Ashish Shenoy and Ayan Chakrabarti and Bilal Piot and Bo Wu and Bobak Shahriari and Bryce Petrini and Charlie Chen and Charline Le Lan and Christopher A. Choquette-Choo and CJ Carey and Cormac Brick and Daniel Deutsch and Danielle Eisenbud and Dee Cattle and Derek Cheng and Dimitris Paparas and Divyashree Shivakumar Sreepathihalli and Doug Reid and Dustin Tran and Dustin Zelle and Eric Noland and Erwin Huizenga and Eugene Kharitonov and Frederick Liu and Gagik Amirkhanyan and Glenn Cameron and Hadi Hashemi and Hanna Klimczak-Plucińska and Harman Singh and Harsh Mehta and Harshal Tushar Lehri and Hussein Hazimeh and Ian Ballantyne and Idan Szpektor and Ivan Nardini and Jean Pouget-Abadie and Jetha Chan and Joe Stanton and John Wieting and Jonathan Lai and Jordi Orbay and Joseph Fernandez and Josh Newlan and Ju-yeong Ji and Jyotinder Singh and Kat Black and Kathy Yu and Kevin Hui and Kiran Vodrahalli and Klaus Greff and Linhai Qiu and Marcella Valentine and Marina Coelho and Marvin Ritter and Matt Hoffman and Matthew Watson and Mayank Chaturvedi and Michael Moynihan and Min Ma and Nabila Babar and Natasha Noy and Nathan Byrd and Nick Roy and Nikola Momchev and Nilay Chauhan and Noveen Sachdeva and Oskar Bunyan and Pankil Botarda and Paul Caron and Paul Kishan Rubenstein and Phil Culliton and Philipp Schmid and Pier Giuseppe Sessa and Pingmei Xu and Piotr Stanczyk and Pouya Tafti and Rakesh Shivanna and Renjie Wu and Renke Pan and Reza Rokni and Rob Willoughby and Rohith Vallu and Ryan Mullins and Sammy Jerome and Sara Smoot and Sertan Girgin and Shariq Iqbal and Shashir Reddy and Shruti Sheth and Siim Põder and Sijal Bhatnagar and Sindhu Raghuram Panyam and Sivan Eiger and Susan Zhang and Tianqi Liu and Trevor Yacovone and Tyler Liechty and Uday Kalra and Utku Evci and Vedant Misra and Vincent Roseberry and Vlad Feinberg and Vlad Kolesnikov and Woohyun Han and Woosuk Kwon and Xi Chen and Yinlam Chow and Yuvein Zhu and Zichuan Wei and Zoltan Egyed and Victor Cotruta and Minh Giang and Phoebe Kirk and Anand Rao and Kat Black and Nabila Babar and Jessica Lo and Erica Moreira and Luiz Gustavo Martins and Omar Sanseviero and Lucas Gonzalez and Zach Gleicher and Tris Warkentin and Vahab Mirrokni and Evan Senter and Eli Collins and Joelle Barral and Zoubin Ghahramani and Raia Hadsell and Yossi Matias and D. Sculley and Slav Petrov and Noah Fiedel and Noam Shazeer and Oriol Vinyals and Jeff Dean and Demis Hassabis and Koray Kavukcuoglu and Clement Farabet and Elena Buchatskaya and Jean-Baptiste Alayrac and Rohan Anil and Dmitry and Lepikhin and Sebastian Borgeaud and Olivier Bachem and Armand Joulin and Alek Andreev and Cassidy Hardin and Robert Dadashi and Léonard Hussenot},
      year={2025},
      eprint={2503.19786},
      archivePrefix={arXiv},
      primaryClass={cs.CL},
      url={https://arxiv.org/abs/2503.19786}, 
}

@misc{gemmateam2026gemma4technicalreport,
      title={Gemma 4 Technical Report}, 
      author={{Gemma Team} and Sherif El Abd and Vaibhav Aggarwal and Robin Algayres and Alek Andreev and Olivier Bachem and Ian Ballantyne and Cormac Brick and Victor Cărbune and Michelle Casbon and Mayank Chaturvedi and Aditya Chawla and Victor Cotruta and Alice Coucke and Phil Culliton and Robert Dadashi and Lucas Dixon and Mohamed Elhawaty and Utku Evci and Clément Farabet and Johan Ferret and Filippo Galgani and Sertan Girgin and Jean-Bastien Grill and Maarten Grootendorst and Jiaxian Guo and Cassidy Hardin and Yanzhang He and Steven M. Hernandez and Omri Homburger and Léonard Hussenot and Juyeong Ji and Armand Joulin and Aishwarya Kamath and Parnian Kassraie and Olivier Lacombe and Preethi Lahoti and Gaël Liu and Gus Martins and Luciano Martins and Tatiana Matejovicova and Ramona Merhej and Nikola Momchev and Sneha Mondal and Ryan Mullins and Sindhu Raghuram Panyam and Shreya Pathak and Sarah Perrin and André Susano Pinto and Etienne Pot and Angéline Pouget and Alexandre Ramé and Sabela Ramos and Douglas Reid and David Rim and Morgane Rivière and Karsten Roth and Louis Rouillard and Omar Sanseviero and Pier Giuseppe Sessa and Shane Settle and Danila Sinopalnikov and Sara Smoot and Piotr Stanczyk and Andreas Steiner and Lawrence Stewart and Ilya Tolstikhin and Michael Tschannen and Anton Tsitsulin and Nino Vieillard and Renjie Wu and Pingmei Xu and Haichuan Yang and Edouard Yvinec and Biao Zhang and Li Zhang and Joe Zou and Nicolas Aagnes and Abdelrahman Abdelhamed and Jakub Adamek and Shivani Agrawal and Shubham Agrawal and Ibrahim Alabdulmohsin and Jean Baptiste Alayrac and Uri Alon and Chandramouli Amarnath and Ankesh Anand and Chrysovalantis Anastasiou and Setareh Ariafar and François-Xavier Aubet and Kyriakos Axiotis and Federico Barbero and Joelle Barral and Alexei Bendebury and Urs Bergmann and Stanley Bileschi and Kat Black and Mathieu Blondel and Sebastian Borgeaud and Arthur Bražinskas and Ryan Burnell and Robert Busa-Fekete and Mu Cai and Daniele Calandriello and Glenn Cameron and Charlotte Caucheteux and Rahma Chaabouni and Garima Chadha and Jetha Chan and Blake Jianhang Chen and Jesse Chen and Lin Chen and Xu Chen and Derek Cheng and Tzu-hsiang Chien and Nikolai Chinaev and Yi Chou and Zhaohui Chu and Benjamin Coleman and Pooja Consul and Sam Conway-Rahman and Scott Crowell and Dylan Cutler and Vivek Dani and Samira Daruki and Anil Das and Daniel Deutsch and Nishanth Dikkala and Li Ding and Qiuhan Ding and Shenil Dodhia and Konstantin Donhauser and Tulsee Doshi and Anca Dragan and Alex Druinsky and Sahil Dua and Zoltan Egyed and Danielle Eisenbud and Daniel Eppens and Cindy Fan and Bahare Fatemi and Yassir Fathullah and Vlad Feinberg and Milen Ferev and Sebastian Flennerhag and Takumi Fujimoto and João Gabriel Oliveira and Isaac Galatzer-Levy and João Gante and Simon Geisler and Soham Ghosal and Antonious M. Girgis and Tamara von Glehn and Alec Go and Alhaad Gokhale and Alex Grills and Yiming Gu and Mayank Gupta and Pramod Gupta and Guru Guruganesh and Raia Hadsell and Hamza Harkous and Jitendra Harlalka and Demis Hassabis and Anja Hauth and Joe Heyward and Arian Hosseini and Chih-Yang Hsia and I-Hung Hsu and Xiaopeng Huang and Yangsibo Huang and Kevin Hui and Adrian Hutter and Te I and Fotis Iliopoulos and Advait Jain and Ganesh Jawahar and Ziwei Ji and Qilin Jin and Melvin Johnson and Kandarp Joshi and Arun Kandoor and Wang-Cheng Kang and Koray Kavukcuoglu and Mehran Kazemi and Kathleen Kenealy and Amr Khalifa and Phoebe Kirk and Ivan Korotkov and Suraj Kothawade and Vitaly Kovalev and Neel Kovelamudi and Adam Kraft and Ravin Kumar and Vivek Kumar and Harish Kuppam and Justin Lannin and Chen-Yu Lee and Seungji Lee and Dmitry Lepikhin and Alon Levkovitch and Dongdong Li and Qiujia Li and Valentin Liévin and Ethan Lin and Ziqian Lin and Casper Liu and Tianlin Liu and Tianqi Liu and Xin Liu and Ivan Lobov and Mayank Lunayach and Min Ma and Gagan Madan and Andrii Maksai and Eric Malmi and Michal Matuszak and Daniel McDuff and Gaurav Menghani and Maciej Mikuła and Daniil Mirylenka and Karolis Misiunas and Vedant Misra and Andreea Mitran and Kareem Mohamed and Maksim Mukha and Eric Noland and James O'Donnell and Brendan O'Donoghue and Kate Olszewska and Bernett Orlando and Wanqiong Pan and Rina Panigrahy and Unnati Parekh and Nicolas Perez-Nieves and Chunjong Park and Eric Paskie and Liqian Peng and Bryce Petrini and Slav Petrov and Jonas Pfeiffer and Bilal Piot and Martyna Plomecka and Siim Poder and Octavio Ponce and Arijit Pramanik and David Racz and Anish Rajan and Michelle Ramanovich and Anand Rao and Marvin Ritter and Vitor Rodrigues and Evan Rosen and Mikołaj Rybiński and Noveen Sachdeva and Michaël E. Sander and Rohit Sathyanarayana and Sagar Savla and Samuel Schmidgall and Tal Schuster and George Scrivener and Benoit Seguin and Andrew Sellergren and Aliaksei Severyn and Izhak Shafran and Dhruv Shah and Bobak Shahriari and Yuan Shangguan and Ashish Shenoy and Pradeep Shenoy and Rakesh Shivanna and Pauline Sho and Lucas Spangher and Wojciech Stokowiec and Tim Strother and Yao Su and Yinghao Sun and Mukund Sundararajan and Andrea Tacchetti and Mor Hazan Taege and Pouya Tafti and Jean Tarbouriech and Chetan Tekur and Shantanu Thakoor and Rahul Thapa and Madeleine Traverse and Lenart Treven and Tao Tu and Chien Te Tung and Çağlar Ünlü and Petar Veličković and Malini Pooni Venkat and Sagar Gubbi Venkatesh and Vidya Venkiteswaran and Francesco Visin and Alex Vitvitskyi and Kiran Vodrahalli and Weiyi Wang and Xin Wang and Tris Warkentin and Jan Wassenberg and John Wieting and Cindy Wu and Lechao Xiao and Hao Xu and Yuhui Xu and Fuzhao Xue and Arun Yadav and Jun Yan and Antoine Yang and Lin Yang and Ming-Hsuan Yang and Ziyu Ying and Jae Hyeon Yoo and Morteza Zadimoghaddam and Sajjad Zafar and Fred Zhang and Jiageng Zhang and Jianyi Zhang and Xiaofan Zhang and Chao Zhao and David Zhou and Chen Zou},
      year={2026},
      eprint={2607.02770},
      archivePrefix={arXiv},
      primaryClass={cs.CL},
      url={https://arxiv.org/abs/2607.02770}, 
}

@article{spearman1904general,
  author  = {Spearman, Charles},
  title   = {The Proof and Measurement of Association between Two Things},
  journal = {The American Journal of Psychology},
  volume  = {15},
  number  = {1},
  pages   = {72--101},
  year    = {1904},
  doi     = {10.2307/1412159}
}

@article{Pearson1895,
    author  = {Pearson, Karl},
    title   = {VII. Note on regression and inheritance in the case of two parents},
    journal = {Proceedings of the Royal Society of London},
    volume  = {58},
    number  = {347--352},
    pages   = {240--242},
    year    = {1895},
    month   = dec,
    doi     = {10.1098/rspl.1895.0041}
}

@article{cai_tsne,
  author  = {T. Tony Cai and Rong Ma},
  title   = {Theoretical Foundations of t-SNE for Visualizing High-Dimensional Clustered Data},
  journal = {Journal of Machine Learning Research},
  year    = {2022},
  volume  = {23},
  number  = {301},
  pages   = {1--54},
  url     = {http://jmlr.org/papers/v23/21-0524.html}
}

@inproceedings{xgb_chen_2016,
author = {Chen, Tianqi and Guestrin, Carlos},
title = {XGBoost: A Scalable Tree Boosting System},
year = {2016},
isbn = {9781450342322},
publisher = {Association for Computing Machinery},
address = {New York, NY, USA},
url = {https://doi.org/10.1145/2939672.2939785},
doi = {10.1145/2939672.2939785},
pages = {785–794},
numpages = {10},
location = {San Francisco, California, USA},
series = {KDD '16}
}

@inproceedings{tyo2023valla,
 title = "Valla: Standardizing and Benchmarking Authorship Attribution and Verification Through Empirical Evaluation and Comparative Analysis",
    author = "Tyo, Jacob  and
      Dhingra, Bhuwan  and
      Lipton, Zachary C.",
    editor = "Park, Jong C.  and
      Arase, Yuki  and
      Hu, Baotian  and
      Lu, Wei  and
      Wijaya, Derry  and
      Purwarianti, Ayu  and
      Krisnadhi, Adila Alfa",
    booktitle = "Proceedings of the 13th International Joint Conference on Natural Language Processing and the 3rd Conference of the Asia-Pacific Chapter of the Association for Computational Linguistics (Volume 1: Long Papers)",
    month = nov,
    year = "2023",
    address = "Nusa Dua, Bali",
    publisher = "Association for Computational Linguistics",
    url = "https://aclanthology.org/2023.ijcnlp-main.43/",
    doi = "10.18653/v1/2023.ijcnlp-main.43",
    pages = "649--660"
}

@inproceedings{boenninghoff-etal-2024-wrote,
    title = "Who Wrote When? Author Diarization in Social Media Discussions",
    author = "Boenninghoff, Benedikt  and
      Hosseini, Henry  and
      Nickel, Robert M.  and
      Kolossa, Dorothea",
    editor = "Al-Onaizan, Yaser  and
      Bansal, Mohit  and
      Chen, Yun-Nung",
    booktitle = "Findings of the Association for Computational Linguistics: EMNLP 2024",
    month = nov,
    year = "2024",
    address = "Miami, Florida, USA",
    publisher = "Association for Computational Linguistics",
    url = "https://aclanthology.org/2024.findings-emnlp.922/",
    doi = "10.18653/v1/2024.findings-emnlp.922",
    pages = "15721--15734",
}

@article{grieve2007quantitative,
  title={Quantitative authorship attribution: An evaluation of techniques},
  author={Grieve, Jack},
  journal={Literary and linguistic computing},
  volume={22},
  number={3},
  pages={251--270},
  year={2007},
  publisher={Oxford University Press},
  doi={https://doi.org/10.1093/llc/fqm020}
}

@inproceedings{zeng2025residualized,
   title = "Residualized Similarity for Faithfully Explainable Authorship Verification",
    author = "Zeng, Peter  and
      Alipoormolabashi, Pegah  and
      Mun, Jihu  and
      Dey, Gourab  and
      Soni, Nikita  and
      Balasubramanian, Niranjan  and
      Rambow, Owen  and
      Schwartz, H.",
    editor = "Christodoulopoulos, Christos  and
      Chakraborty, Tanmoy  and
      Rose, Carolyn  and
      Peng, Violet",
    booktitle = "Findings of the Association for Computational Linguistics: EMNLP 2025",
    month = nov,
    year = "2025",
    address = "Suzhou, China",
    publisher = "Association for Computational Linguistics",
    url = "https://aclanthology.org/2025.findings-emnlp.856/",
    doi = "10.18653/v1/2025.findings-emnlp.856",
    pages = "15824--15837",
    ISBN = "979-8-89176-335-7",
}

@inproceedings{gupta2019authorship,
 author = {Gupta, Shriya TP and Sahoo, Jajati Keshari and Roul, Rajendra Kumar},
title = {Authorship Identification using Recurrent Neural Networks},
year = {2019},
isbn = {9781450366359},
publisher = {Association for Computing Machinery},
address = {New York, NY, USA},
url = {https://doi.org/10.1145/3325917.3325935},
doi = {10.1145/3325917.3325935},
booktitle = {Proceedings of the 2019 3rd International Conference on Information System and Data Mining},
pages = {133–137},
numpages = {5},
location = {Houston, TX, USA},
series = {ICISDM '19}
}

@inproceedings{qian2017deep,
  title={Deep Learning based Authorship Identification},
  author={Chen Qian and Ting He and Zhang-Wen Rao},
  year={2017},
  url={https://api.semanticscholar.org/CorpusID:42982101}
}

@inproceedings{boenninghoff2019explainable,
  title={Explainable authorship verification in social media via attention-based similarity learning},
  author={Boenninghoff, Benedikt and Hessler, Steffen and Kolossa, Dorothea and Nickel, Robert M},
  booktitle={2019 IEEE International Conference on Big Data (Big Data)},
  pages={36--45},
  year={2019},
  organization={IEEE},
  doi={10.1109/BigData47090.2019.9005650}
}

@inproceedings{rivera2021learning,
    title = "Learning Universal Authorship Representations",
    author = "Rivera-Soto, Rafael A.  and
      Miano, Olivia Elizabeth  and
      Ordonez, Juanita  and
      Chen, Barry Y.  and
      Khan, Aleem  and
      Bishop, Marcus  and
      Andrews, Nicholas",
    editor = "Moens, Marie-Francine  and
      Huang, Xuanjing  and
      Specia, Lucia  and
      Yih, Scott Wen-tau",
    booktitle = "Proceedings of the 2021 Conference on Empirical Methods in Natural Language Processing",
    month = nov,
    year = "2021",
    address = "Online and Punta Cana, Dominican Republic",
    publisher = "Association for Computational Linguistics",
    url = "https://aclanthology.org/2021.emnlp-main.70/",
    doi = "10.18653/v1/2021.emnlp-main.70",
    pages = "913--919",
}

@inproceedings{fabien2020bertaa,
 title = "{B}ert{AA} : {BERT} fine-tuning for Authorship Attribution",
    author = {Fabien, Ma{\"e}l  and
      Villatoro-Tello, Esau  and
      Motlicek, Petr  and
      Parida, Shantipriya},
    editor = "Bhattacharyya, Pushpak  and
      Sharma, Dipti Misra  and
      Sangal, Rajeev",
    booktitle = "Proceedings of the 17th International Conference on Natural Language Processing (ICON)",
    month = dec,
    year = "2020",
    address = "Indian Institute of Technology Patna, Patna, India",
    publisher = "NLP Association of India (NLPAI)",
    url = "https://aclanthology.org/2020.icon-main.16/",
    pages = "127--137",
}

@inproceedings{huang2024can,
    title = "Can Large Language Models Identify Authorship?",
    author = "Huang, Baixiang  and
      Chen, Canyu  and
      Shu, Kai",
    editor = "Al-Onaizan, Yaser  and
      Bansal, Mohit  and
      Chen, Yun-Nung",
    booktitle = "Findings of the {A}ssociation for {C}omputational {L}inguistics: {EMNLP} 2024",
    month = nov,
    year = "2024",
    address = "Miami, Florida, USA",
    publisher = "Association for Computational Linguistics",
    url = "https://aclanthology.org/2024.findings-emnlp.26/",
    doi = "10.18653/v1/2024.findings-emnlp.26",
    pages = "445--460",
}

@inproceedings{ramnath2025cave,
   title = "{CAVE}: Controllable Authorship Verification Explanations",
    author = "Ramnath, Sahana  and
      Pandey, Kartik  and
      Boschee, Elizabeth  and
      Ren, Xiang",
    editor = "Chiruzzo, Luis  and
      Ritter, Alan  and
      Wang, Lu",
    booktitle = "Proceedings of the 2025 Conference of the Nations of the Americas Chapter of the Association for Computational Linguistics: Human Language Technologies (Volume 1: Long Papers)",
    month = apr,
    year = "2025",
    address = "Albuquerque, New Mexico",
    publisher = "Association for Computational Linguistics",
    url = "https://aclanthology.org/2025.naacl-long.451/",
    doi = "10.18653/v1/2025.naacl-long.451",
    pages = "8939--8961",
    ISBN = "979-8-89176-189-6",
}

@techreport{openai_gpt-4,
  title        = {GPT-4},
  author       = {{OpenAI}},
  institution  = {OpenAI},
  year         = {2023},
  url          = {https://cdn.openai.com/papers/gpt-4-system-card.pdf}
}

@techreport{openai_gpt-3.5,
  author       = {{OpenAI}},
  title        = {GPT-3.5},
  year         = {2022},
  institution  = {OpenAI},
  url = {https://platform.openai.com/docs/models/gpt-3.5-turbo},
}

@inproceedings{qiu-etal-2025-mstyledistance,
    title = "m{S}tyle{D}istance: Multilingual Style Embeddings and their Evaluation",
    author = "Qiu, Justin  and
      Zhu, Jiacheng  and
      Patel, Ajay  and
      Apidianaki, Marianna  and
      Callison-Burch, Chris",
    editor = "Che, Wanxiang  and
      Nabende, Joyce  and
      Shutova, Ekaterina  and
      Pilehvar, Mohammad Taher",
    booktitle = "Findings of the Association for Computational Linguistics: ACL 2025",
    month = jul,
    year = "2025",
    address = "Vienna, Austria",
    publisher = "Association for Computational Linguistics",
    url = "https://aclanthology.org/2025.findings-acl.869/",
    doi = "10.18653/v1/2025.findings-acl.869",
    pages = "16917--16931",
    ISBN = "979-8-89176-256-5"
}

@inproceedings{murauer2019generating,
author="Murauer, Benjamin
and Specht, G{\"u}nther",
editor="Crestani, Fabio
and Braschler, Martin
and Savoy, Jacques
and Rauber, Andreas
and M{\"u}ller, Henning
and Losada, David E.
and Heinatz B{\"u}rki, Gundula
and Cappellato, Linda
and Ferro, Nicola",
title="Generating Cross-Domain Text Classification Corpora from Social Media Comments",
booktitle="Experimental IR Meets Multilinguality, Multimodality, and Interaction",
year="2019",
publisher="Springer International Publishing",
address="Cham",
pages="114--125",
isbn="978-3-030-28577-7",
doi = {https://doi.org/10.1007/978-3-030-28577-7_7}
}

@article{halvani2016authorship,
title = {Authorship verification for different languages, genres and topics},
journal = {Digital Investigation},
volume = {16},
pages = {S33-S43},
year = {2016},
note = {DFRWS 2016 Europe},
issn = {1742-2876},
doi = {https://doi.org/10.1016/j.diin.2016.01.006},
url = {https://www.sciencedirect.com/science/article/pii/S1742287616000074},
author = {Oren Halvani and Christian Winter and Anika Pflug},
}

@inproceedings{
hu2022lora,
title={Lo{RA}: Low-Rank Adaptation of Large Language Models},
author={Edward J Hu and Yelong Shen and Phillip Wallis and Zeyuan Allen-Zhu and Yuanzhi Li and Shean Wang and Lu Wang and Weizhu Chen},
booktitle={International Conference on Learning Representations},
year={2022},
url={https://openreview.net/forum?id=nZeVKeeFYf9}
}

@article{youden1950index,
  title={Index for rating diagnostic tests},
  author={Youden, William J},
  journal={Cancer},
  volume={3},
  number={1},
  pages={32--35},
  year={1950},
  publisher={Wiley Online Library},
  doi = {https://doi.org/10.1002/1097-0142(1950)3:1<32::AID-CNCR2820030106>3.0.CO;2-3}
}

@article{efron_bootstrap_1979,
author = {Bradley Efron},
title = {{Bootstrap Methods: Another Look at the Jackknife}},
volume = {7},
journal = {The Annals of Statistics},
number = {1},
publisher = {Institute of Mathematical Statistics},
pages = {1 -- 26},
year = {1979},
doi = {10.1214/aos/1176344552},
URL = {https://doi.org/10.1214/aos/1176344552}
}

@inproceedings{Ansel_PyTorch_2_Faster_2024,
author = {Ansel, Jason and Yang, Edward and He, Horace and Gimelshein, Natalia and Jain, Animesh and Voznesensky, Michael and Bao, Bin and Bell, Peter and Berard, David and Burovski, Evgeni and Chauhan, Geeta and Chourdia, Anjali and Constable, Will and Desmaison, Alban and DeVito, Zachary and Ellison, Elias and Feng, Will and Gong, Jiong and Gschwind, Michael and Hirsh, Brian and Huang, Sherlock and Kalambarkar, Kshiteej and Kirsch, Laurent and Lazos, Michael and Lezcano, Mario and Liang, Yanbo and Liang, Jason and Lu, Yinghai and Luk, CK and Maher, Bert and Pan, Yunjie and Puhrsch, Christian and Reso, Matthias and Saroufim, Mark and Siraichi, Marcos Yukio and Suk, Helen and Suo, Michael and Tillet, Phil and Wang, Eikan and Wang, Xiaodong and Wen, William and Zhang, Shunting and Zhao, Xu and Zhou, Keren and Zou, Richard and Mathews, Ajit and Chanan, Gregory and Wu, Peng and Chintala, Soumith},
booktitle = {29th ACM International Conference on Architectural Support for Programming Languages and Operating Systems, Volume 2 (ASPLOS '24)},
doi = {10.1145/3620665.3640366},
month = apr,
publisher = {ACM},
title = {{PyTorch 2: Faster Machine Learning Through Dynamic Python Bytecode Transformation and Graph Compilation}},
url = {https://docs.pytorch.org/assets/pytorch2-2.pdf},
year = {2024}
}

@software{von_Werra_TRL_Transformers_Reinforcement_2020,
author = {von Werra, Leandro and Belkada, Younes and Tunstall, Lewis and Beeching, Edward and Thrush, Tristan and Lambert, Nathan and Huang, Shengyi and Rasul, Kashif and Gallouédec, Quentin},
license = {Apache-2.0},
month = mar,
title = {{TRL: Transformers Reinforcement Learning}},
url = {https://github.com/huggingface/trl},
version = {1.2},
year = {2020}
}

@inproceedings{Wolf_Transformers_State-of-the-Art_Natural_2020,
author = {Wolf, Thomas and Debut, Lysandre and Sanh, Victor and Chaumond, Julien and Delangue, Clement and Moi, Anthony and Cistac, Perric and Ma, Clara and Jernite, Yacine and Plu, Julien and Xu, Canwen and Le Scao, Teven and Gugger, Sylvain and Drame, Mariama and Lhoest, Quentin and Rush, Alexander M.},
month = oct,
pages = {38--45},
publisher = {Association for Computational Linguistics},
title = {{Transformers: State-of-the-Art Natural Language Processing}},
url = {https://www.aclweb.org/anthology/2020.emnlp-demos.6},
year = {2020}
}

@inproceedings{kwon2023efficient,
  title={Efficient Memory Management for Large Language Model Serving with PagedAttention},
  author={Woosuk Kwon and Zhuohan Li and Siyuan Zhuang and Ying Sheng and Lianmin Zheng and Cody Hao Yu and Joseph E. Gonzalez and Hao Zhang and Ion Stoica},
  booktitle={Proceedings of the ACM SIGOPS 29th Symposium on Operating Systems Principles},
  year={2023},
  doi = {10.1145/3600006.3613165}
}

@misc{hu2024instructav,
      title={InstructAV: Instruction Fine-tuning Large Language Models for Authorship Verification}, 
      author={Yujia Hu and Zhiqiang Hu and Chun-Wei Seah and Roy Ka-Wei Lee},
      year={2024},
      eprint={2407.12882},
      archivePrefix={arXiv},
      primaryClass={cs.CL},
      url={https://arxiv.org/abs/2407.12882}, 
}

@article{eder2015size,
    author = {Eder, Maciej},
    title = {Does size matter? Authorship attribution, small samples, big problem},
    journal = {Digital Scholarship in the Humanities},
    volume = {30},
    number = {2},
    pages = {167-182},
    year = {2015},
    month = {06},
    issn = {2055-7671},
    doi = {10.1093/llc/fqt066},
    url = {https://doi.org/10.1093/llc/fqt066},
    eprint = {https://academic.oup.com/dsh/article-pdf/30/2/167/21517531/fqt066.pdf},
}

@article{Alfassi2026fanfiction,
author = {Alfassi, Roi and Cooper, Angelora and Mitchell, Zoe and Calabro, Mary and Shaer, Orit and Mokryn, Osnat},
title = {Fanfiction in the Age of AI: Community Perspectives on Creativity, Authenticity and Adoption},
journal = {International Journal of Human–Computer Interaction},
volume = {42},
number = {5},
pages = {3062--3094},
year = {2026},
publisher = {Taylor \& Francis},
doi = {10.1080/10447318.2025.2531272},
URL = {https://doi.org/10.1080/10447318.2025.2531272},
eprint = {https://doi.org/10.1080/10447318.2025.2531272}
}

@article{stamatatos2009survey,
  title={A survey of modern authorship attribution methods},
  author={Stamatatos, Efstathios},
  journal={Journal of the American Society for information Science and Technology},
  volume={60},
  number={3},
  pages={538--556},
  year={2009},
  publisher={Wiley Online Library},
  url={http://onlinelibrary.wiley.com/doi/10.1002/asi.21001/full}
}
